\documentclass[conference]{IEEEtran}
\IEEEoverridecommandlockouts
\usepackage{cite}
\usepackage{amsmath,amssymb,amsfonts}
\usepackage{algorithmic}
\usepackage{graphicx}
\usepackage{textcomp}
\usepackage{xcolor}
\usepackage{url}
\def\BibTeX{{\rm B\kern-.05em{\sc i\kern-.025em b}\kern-.08em
    T\kern-.1667em\lower.7ex\hbox{E}\kern-.125emX}}
\begin{document}

\title{EGAD: Evolving Graph Representation Learning\\ with Self-Attention and Knowledge Distillation\\ for Live Video Streaming Events}

\author{\IEEEauthorblockN{Stefanos Antaris}
\IEEEauthorblockA{KTH Royal Institute of Technology \\
HiveStreaming AB\\
Sweden \\
antaris@kth.se}
\and
\IEEEauthorblockN{Dimitrios Rafailidis}
\IEEEauthorblockA{Maastricht University \\
Netherlands \\
dimitrios.rafailidis@maastrichtuniversity.nl}
\and
\IEEEauthorblockN{Sarunas Girdzijauskas}
\IEEEauthorblockA{KTH Royal Institute of Technology \\
Sweden \\
sarunasg@kth.se}
}
\IEEEoverridecommandlockouts
\IEEEpubid{\makebox[\columnwidth]{978-1-7281-6251-5/20/\$31.00~\copyright2020 IEEE \hfill} \hspace{\columnsep}\makebox[\columnwidth]{ }}
\maketitle
\IEEEpubidadjcol

\begin{abstract}
In this study, we present a dynamic graph representation learning model on weighted graphs to accurately predict the network capacity of connections between viewers in a live video streaming event. We propose EGAD, a neural network architecture to capture the graph evolution by introducing a self-attention mechanism on the weights between consecutive graph convolutional networks. In addition, we account for the fact that neural architectures require a huge amount of parameters to train, thus increasing the online inference latency and negatively influencing the user experience in a live video streaming event. To address the problem of the high online inference of a vast number of parameters, we propose a knowledge distillation strategy. In particular, we design a distillation loss function, aiming to first pretrain a teacher model on offline data, and then transfer the knowledge from the teacher to a smaller student model with less parameters. We evaluate our proposed model on the link prediction task on three real-world datasets, generated by live video streaming events. The events lasted 80 minutes and each viewer exploited the distribution solution provided by the company Hive Streaming AB. The experiments demonstrate the effectiveness of the proposed model in terms of link prediction accuracy and number of required parameters, when evaluated against state-of-the-art approaches. In addition, we study the distillation performance of the proposed model in terms of compression ratio for different distillation strategies, where we show that the proposed model can achieve a compression ratio up to 15:100, preserving high link prediction accuracy. For reproduction purposes, our evaluation datasets and implementation are publicly available at \url{https://stefanosantaris.github.io/EGAD}.
\end{abstract}


\section{Introduction}

\begin{figure}[t!]

    \centering
    \includegraphics[scale=0.14]{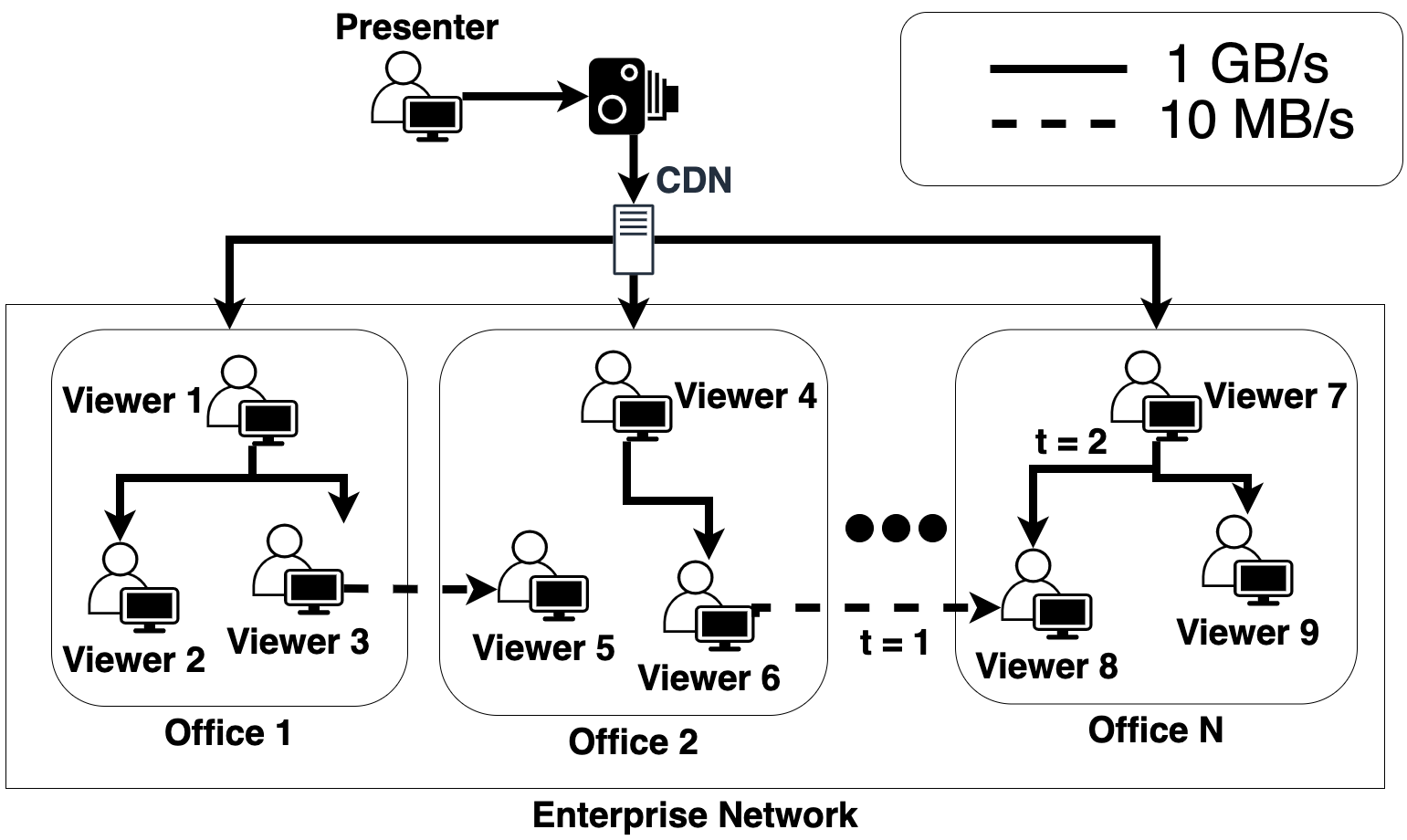}
    \caption{A distributed live video streaming process in enterprise networks.}
    \label{fig:video_distr}
\end{figure}

Nowadays, live video streaming has emerged as a prominent communication solution for several companies worldwide. For example, live video streaming is employed for corporate internal communications, marketing announcements, and so on~\cite{FAN2018332, FARANMAJEED2017103}. Delivering a high quality video to enterprise offices is a challenging task, which stems from the bandwidth requirement, increasing along with the number of viewers in each office. To overcome this challenge, distributed live video streaming solutions were proposed (e.g. by Hive Streaming AB) to deliver high quality video content to several enterprise offices \cite{Roverso2015, Roverso2013}. As shown in Figure \ref{fig:video_distr}, Viewers $1$, $4$ and $7$ download the video content of the presenter directly from the Content Delivery Network (CDN) server. Thereafter, Viewers $1$, $4$ and $7$ have to distribute the video content to the rest of the viewers, that is Viewers $2$, $3$, $5$, $6$, $8$, and $9$. To efficiently distribute the video content, each viewer should establish connections with other viewers of the same office and exploit the internal high-bandwidth network of the office ($1$ GB/s). However, to efficiently establish connections between viewers, Viewer $1$ requires the information that Viewers $2$ and $3$ share the same office. Without this information, Viewer $3$ might erroneously establish a connection to Viewer $5$ of a different office through a low bandwidth network ($10$ MB/s). This will negatively impact the video distribution process of Viewer $5$, as the only established connection of Viewer $5$ will not satisfy the bandwidth requirements of a high quality live video streaming event \cite{Roverso2013}. Nonetheless, this requires the information of the customers' network topology during the live video streaming event, for instance, Viewers $1$, $2$ and $3$ are in Office~$1$. However, it is not always feasible to acquire this information, for example, large enterprises provide limited information about their network topologies for security reasons, or enterprises constantly adapt their networks to assure the desired business outcomes and improve the user experience \cite{Phan2019, Rivera2019}. In addition, complying with the recent data protection regulations (GDPR) \cite{gdpr}, live video streaming providers, such as Hive Streaming AB, are prohibited to retrieve certain network characteristics, such as private and public internet protocol (IP) addresses. Therefore, it is important to predict the network capacity of each connection - bandwidth during a live video streaming event, based on the limited information provided by the already established connections. In doing so, we can infer if the viewers are located in the same office so as to establish connection through the internal high bandwidth network.

\begin{figure*}[th] \centering
\begin{tabular}{ccc}
\includegraphics[scale=0.2]{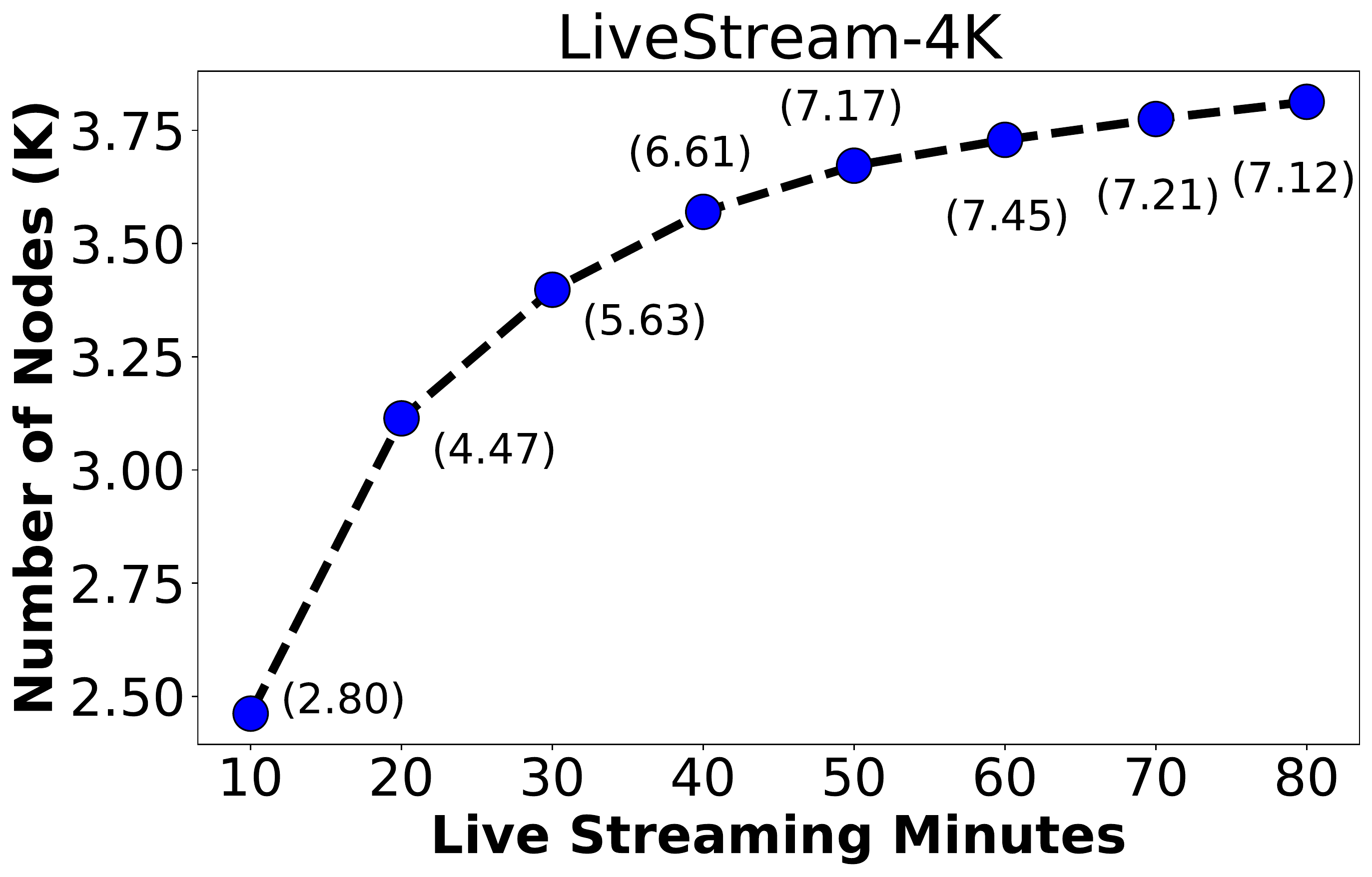} &
\includegraphics[scale=0.2]{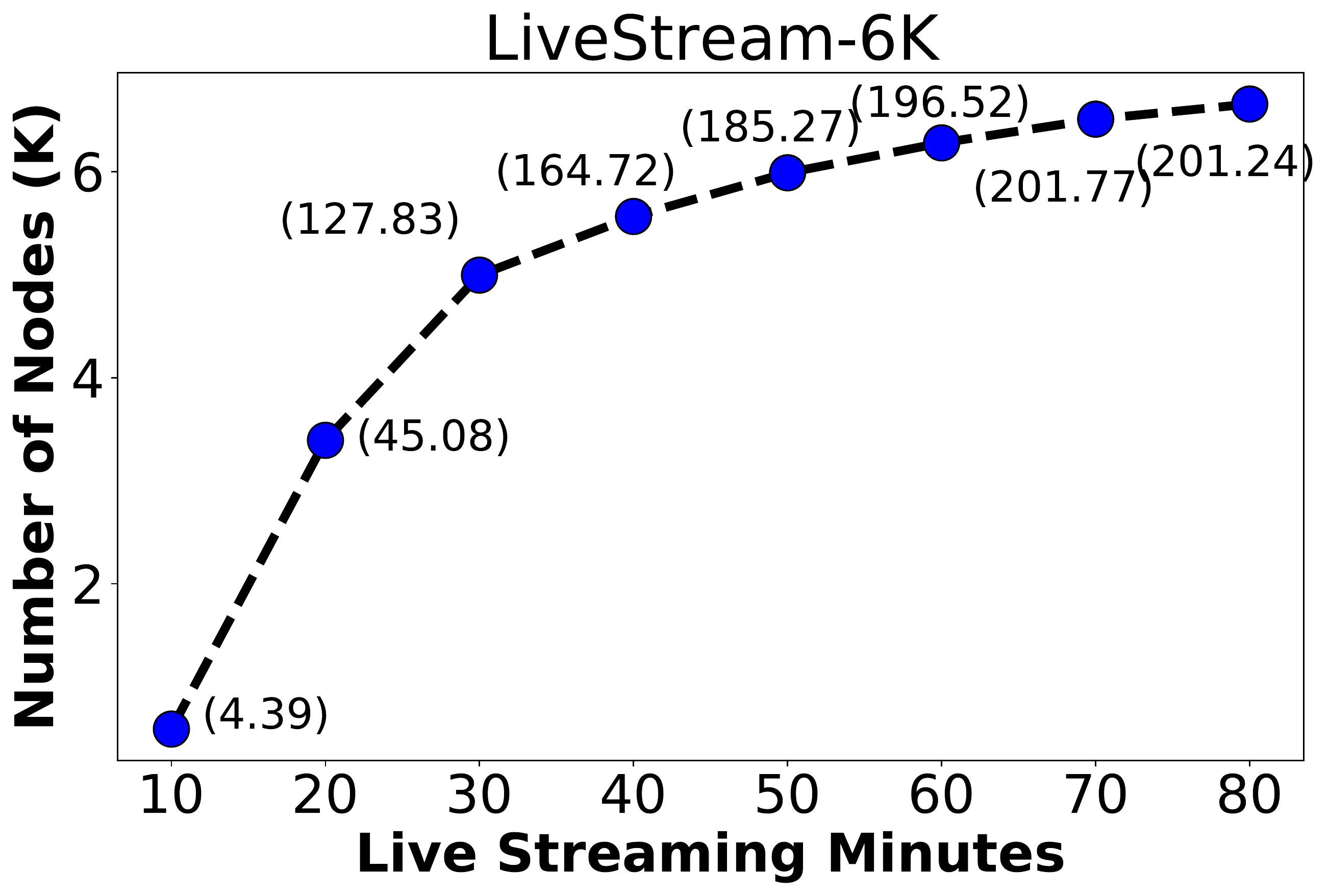} &
\includegraphics[scale=0.2]{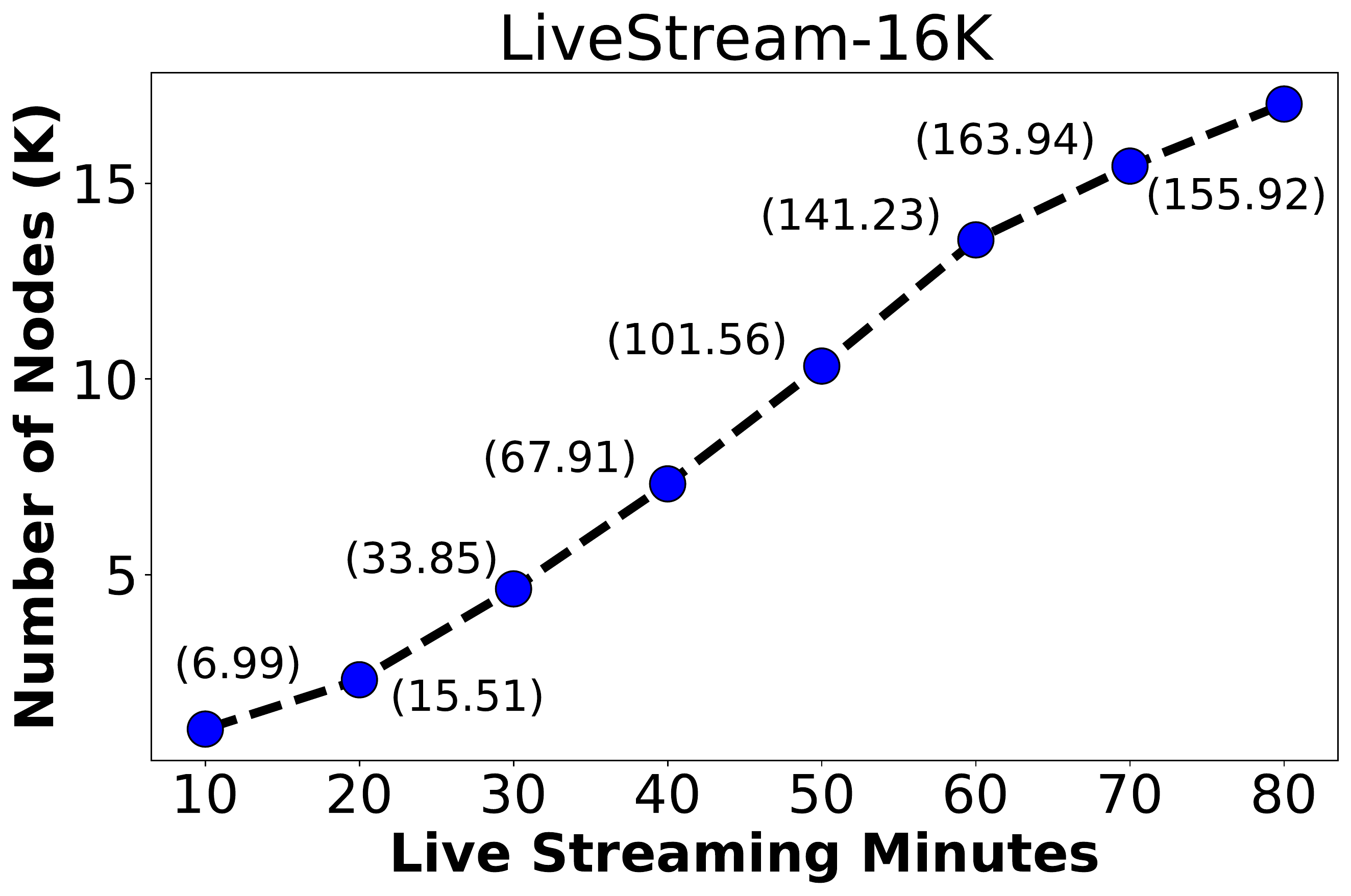} 

\end{tabular}

\vspace{-0.2cm}
\caption{Number of nodes/viewers during the three live video streaming events. In each parenthesis, we denote the respective number of connections/edges (K) among the viewers at a certain snapshot.} \label{fig:NodeEvolution}
\end{figure*}

During a live video streaming event, each viewer has a limited number of connections. In addition, the viewers adapt their connections in real-time so as to improve the distribution of the video content \cite{Roverso2013}. For example, in Figure \ref{fig:video_distr}, Viewer $6$ and $8$ are connected with a low bandwidth network at time step $t=1$. As a consequence, Viewer $8$ drops the connection with Viewer $6$ at time step $t=2$ to establish a connection with Viewer $7$ via a high bandwidth network connection. The effectiveness of a distributed live video streaming solution depends on the accuracy of each viewers' predictions, that is to predict the connections between viewers in the same office. Moreover, the predictions of the viewers' connections have to be performed in a nearly real-time computational time, otherwise it will negatively impact the user experience during the live video streaming event. In this study, we model an enterprise live video streaming event as a dynamic undirected and weighted graph, where the edges weight correspond to the throughput of the connection between two nodes/viewers. The graph nodes/viewers emerge and leave at unexpected rate and each node/viewer adapts their edges/connections, so as to identify the nodes/viewers that are located in the same office and efficiently distribute the video content. Provided that an enterprise live video streaming event has thousands of viewers, such graphs are highly dimensional and sparse. 

\textbf{Graph representation learning.} Recently, graph representation learning approaches emerged that compute compact latent node representations to solve the graph dimensionality problem \cite{Grover2016,hamilton17,Perozzi2014}. Calculating the latent node/viewer representations has proven a successful means to address the link prediction problem on graphs \cite{Grover2016, Sankar2020, Pareja2020, kipf2016variational, mahdavi2019, hamilton2017}. Baseline graph representation learning approaches exploit random walks to learn the latent node/viewer representations \cite{Grover2016, Perozzi2014}. More recently, several studies design different neural network architectures to calculate complex patterns in graph structures \cite{kipf2016variational, velickovic2018, hamilton2017, Cao2015, Wang16, Kefato2020}. However, these neural network architectures work on static graphs. To capture the graph evolution, recent approaches employ Recurrent Neural Networks (RNN) \cite{Pareja2020,Goyal2020} and self-attention mechanisms \cite{Sankar2020} between consecutive graph snapshots. Although dynamic graph representation learning approaches achieve high accuracy in link prediction, the underlying neural networks require to train a large amount of parameters. Therefore, these approaches incur high latency during the online inference of the node/viewer representations due to the large model sizes of the underlying neural network architectures \cite{Jiaxi2018, Jimmy2014, hinton2015,Liu_2019_CVPR,Wang2019, YimJBK17}. As a consequence, state-of-the-art approaches are not applicable to real-world live video streaming solutions, as the high online latency inference increases the computational time of link prediction during a live video streaming event, resulting in high complexity when adapting the viewers' connections.

\textbf{Knowledge distillation.} Alternatively, to reduce the high online latency inference, graph representation learning approaches could employ neural networks of smaller sizes with less parameters. However, such models might fail to accurately capture the structure of an evolving graph, resulting in low link prediction accuracy. \textit{Knowledge distillation} has been recently introduced as a model-independent strategy to generate a small model that exhibits low online latency inference, while preserving high accuracy \cite{hinton2015, Bucilua2006, Jiaxi2018}. The main idea of knowledge distillation is to train a large model, namely \textit{teacher}, as an offline training process. The teacher model is a neural network architecture that requires to train a large number of parameters, so as to learn the structure of offline data. Having pretrained the teacher model, the knowledge distillation strategies compute a smaller \textit{student} model with less parameters, that is more suitable for deployment in production. In particular, the student model is trained on online data, and distills the knowledge of the pretrained teacher model. This means that the student model mimics the teacher model and preserves the high prediction accuracy, while at the same time reduces the online inference of the model parameters due to its small size \cite{Mary2019, Bucilua2006}. A few attempts have been made on graph representation learning with knowledge distillation strategies to reduce the model sizes of the underlying neural network architectures \cite{Jiaqi2019, Lassance2019, LeeS19}. As we will show in Section \ref{sec:distil}, such approaches fail to achieve a high compression ratio on the student model, that is the size of the student model remains high when compared with the size of the teacher model. This occurs because these approaches learn low dimensional representations on static graphs, which do not correspond to the dynamic case of live video streaming events.

\textbf{Contribution.} To overcome the limitations of existing models, in this work we present a knowledge distillation strategy for dynamic graph representation learning, namely EGAD, for the link prediction task during live video streaming events. Our main contributions are summarized as follows: 
\begin{itemize}
    \item EGAD employs a self-attention mechanism on the weights of consecutive Graph Convolutional Networks (GCNs), to capture the graph evolution and learn accurate latent node/viewer representations, during a live video streaming event.
    \item To the best of our knowledge we are the first to study knowledge distillation for dynamic graph representation learning. We train the EGAD teacher model in an offline process and formulate a distillation loss function to transfer the pretrained knowledge to a smaller student model on online data. In doing so, we significantly reduce the number of parameters when training the student model on online data, and achieve high link prediction accuracy.
\end{itemize}
Our experiments on real-world datasets of live video streaming events demonstrate the superiority of the proposed model to accurately capture the evolution of the graph and reduce the online latency inference of the model parameters, when compared with other state-of-the-art methods.

The remainder of the paper is organized as follows: in Section~\ref{sec:data_analysis} we present the collected live video streaming data in Hive Streaming AB, and in Section \ref{sec:prop_method} we detail the proposed model. Our experimental evaluation is presented in Section \ref{sec:exp_eval}, and we conclude the study in Section \ref{sec:conc}.

\section{Live Video Streaming Data} \label{sec:data_analysis}

During a live video streaming event in Hive Streaming AB, various data are collected such as connections per viewer, throughput per connection, and so on, to provide valuable insights to customers. Each viewer periodically reports the data to centralized servers. To evaluate the performance of the proposed model, we collected real-world datasets based on the reports of three live video streaming events, that is LiveStream-4K, LiveStream-6K and LiveStream-16K. All datasets are anonymized and publicly available. The duration of each live video streaming event is $80$ minutes. Each generated dataset consists of $8$ weighted undirected graph/viewing snapshots, corresponding to the viewers' connections every $10$ minutes. A weight of a graph/viewing edge corresponds to the throughput of the connection among two viewers at each snapshot. The LiveStream-4K dataset has $3,813$ viewers, distributed  to $15$ different offices, and $11,066$ connections. In the LiveStream-6K dataset, $6,655$ viewers attended the live video streaming event from $29$ different offices. The viewers established $787,291$ connections. The LiveStream-16K dataset consists of $17,026$ viewers and $482,185$ connections in total. The viewers participated in the live video streaming event from $46$ different offices.

Figure \ref{fig:NodeEvolution} illustrates the different patterns of how viewers emerge during the three live video streaming events. LiveStream-4K has more viewers than LiveStream-6K and LiveStream-16K, during the first $10$ minutes of the live video streaming event. This indicates that in LiveStream-4K the majority of the viewers started to attend the live video streaming event from the beginning. In LiveStream-6K, the first $2$ graph/viewing snapshots significantly change in terms of number of viewers, for $0-20$ minutes $2.8$K new viewers emerged, whereas in LiveStream-4K and LiveStream-16K $0.5$K and $1$K viewers emerged, respectively. LiveStream-4K is less informative as the viewers establish the lowest number of connections. Finally, we can observe that viewers in LiveStream-16K emerge at the lowest pace during the live video streaming event. As we will demonstrate in Section \ref{sec:perf_eval}, the effectiveness of the proposed knowledge distillation strategy and baseline approaches not only depends on the graph sizes but also on different patterns that viewers emerge during the live video streaming events.

\section{Proposed Method} \label{sec:prop_method}

A live video streaming event is represented as a sequence of $K$ graph/viewing snapshots $\mathcal{G} = \{\mathcal{G}_1, \ldots, \mathcal{G}_K\}$. $\forall$ $k=1,\ldots,K$ snapshot we consider the graph $\mathcal{G}_k = (\mathcal{V}_k, \mathcal{E}_k, \mathbf{X}_k)$, where $\mathcal{V}_k$ corresponds to the set of $n_k=|\mathcal{V}_k|$ viewers, $\mathcal{E}_k$ is the set of connections, and $\mathbf{X}_k \in \mathbb{R}^{n_k\times m}$ is the matrix of the $m$ features of each viewer. For each graph $\mathcal{G}_k$, we consider a weighted adjacency matrix $\mathbf{A}_k \in \mathbb{R}^{n_k \times n_k}$, where $A(u,v) > 0$ for the viewers $u \in \mathcal{V}_k$ and $v \in \mathcal{V}_k$, if $e_k(u,v) \in \mathcal{E}_k$. The weight $A(u,v)$ corresponds to the bandwidth measured between viewers $u \in \mathcal{V}_k$ and $v \in \mathcal{V}_k$ at the $k$-th snapshot. Given a sequence of $l$ graph/viewing snapshots\footnote{The reason for not accounting for all the previous snapshots from the beginning of the live video streaming event, and consider only a certain time window $l$ is because we observed in our experiments that large values of $l$ do not necessarily increase the prediction accuracy, while at the same time significantly increase the number of the model parameters. The influence of $l$ on the performance of the proposed model and the baseline approaches is studied in Table IV.} $\{\mathcal{G}_{k-l},\ldots,\mathcal{G}_k\}$, the goal of the proposed model is to compute $d$-dimensional latent representations $\mathbf{Z}_k \in \mathbb{R}^{n_k \times d}$, with $d \ll m$ \cite{Sankar2020, mahdavi2019, Pareja2020}. The constructed latent representations should capture both the structure of the graph at the graph/viewing snapshot $k$ and the evolutionary behavior of the viewers up to the $k$-th minute. 

Dynamic graph representation learning models employ deep neural network architectures, requiring to train a large amount of parameters \cite{Pareja2020, Zhou2018, Goyal2018}. Such models are computationally expensive to deploy to a large number of viewers in live video streaming events as they incur significant online latency to calculate the viewers' representations \cite{hinton2015,Jimmy2014, Jiaqi2019, Lassance2019}. The problem of knowledge distillation is to generate a smaller online \textit{student} model $\mathcal{S}$ than a pretrained offline large teacher model $\mathcal{T}$. The goal is to reduce the number of trainable parameters of the student model $\mathcal{S}$ to minimize the online latency inference \cite{Bucilua2006, anil2018}. In practice, the teacher model $\mathcal{T}$ is pretrained using a computationally expensive deep neural network architecture to calculate the latent representations $\mathbf{Z}_k^{\mathcal{T}}$ of the offline data. Having trained the teacher model offline, the student model $\mathcal{S}$ learns the latent representations $\mathbf{Z}_k^{\mathcal{S}}$ by minimizing a distillation loss function $L^{\mathcal{D}}$. The distillation loss function $L^{\mathcal{D}}$ calculates the prediction error of the student model $\mathcal{S}$ and the deviation from the latent representations $\mathbf{Z}_k^{\mathcal{T}}$ generated by the teacher model $\mathcal{T}$. This means that the student model $\mathcal{S}$ is able to mimic the already pretrained teacher model $\mathcal{T}$ with fewer parameters \cite{Mary2019, Bucilua2006}. In Section \ref{sec:egad_teacher} we present the offline teacher model EGAD-$\mathcal{T}$, and then in Section \ref{sec:egad_student} we describe the distillation process of the online student model EGAD-$\mathcal{S}$.

\subsection{EGAD-$\mathcal{T}$ Teacher Model} \label{sec:egad_teacher}

The teacher model EGAD-$\mathcal{T}$ learns the viewer representations $\mathbf{Z}_k^{\mathcal{T}}$ at the $k$-th graph/viewing snapshot using $l$ consecutive Graph Convolutional Network (GCN) models \cite{kipf2017, Pareja2020, Seo2016}, with EGAD-$\mathcal{T} = \{GCN_{k-l}, \ldots\, GCN_k\}$, and $l$ being the number of previous graph/viewing snapshots. The input of each $GCN_k$ model is the normalized adjacency matrix $\mathbf{\hat{A}}_k \in \mathbb{R}^{n_k \times n_k}$ and the viewers' features $\mathbf{X}_k$. Provided that the graphs during the live video streaming events have nodes with no features, the node feature matrix $\mathbf{X}_k$ is replaced by the identity matrix $\mathbf{I} \in \mathbb{R}^{n \times n}$, with $m=n$. Each $GCN_k$ model calculates the viewers representations $\mathbf{Z}_k^{\mathcal{T}}$ by applying two convolution layers to $\mathbf{\hat{A}}_k$ and $\mathbf{X}_k$, as follows:

\begin{equation}
    \mathbf{Z}_k^{\mathcal{T}} = \hat{\mathbf{A}}_k ReLU (\hat{\mathbf{A}}_k \mathbf{X}_k \mathbf{W}_k^1) \mathbf{W}_k^2
    \label{eq:gcn}
\end{equation}
where $\mathbf{W}_k^i \in \mathbb{R}^{d_{i-1} \times d_i}$ is the weight parameter matrix of the $i$-th convolutional layer, with $d_i < d_{i-1} < m$. Following~\cite{kipf2016variational, kipf2017} we employ two convolutional layers ($i=1,2$), to learn the weight parameter matrices $\mathbf{W}_k^1 \in \mathbb{R}^{m \times d_1}$ and $\mathbf{W}_k^2 \in \mathbb{R}^{d_1 \times d_2}$, with $d_2 = d$, so as to compute the $d$-dimensional representations $\mathbf{Z}_k^{\mathcal{T}}$. The symmetrically normalized adjacency matrix $\hat{\mathbf{A}}_k$ is calculated as follows: 

\begin{equation}
    \begin{array}{l}
         \hat{\mathbf{A}}_k = \mathbf{D}_k^{-\frac{1}{2}} \tilde{\mathbf{A}}_k \mathbf{D}_k^{-\frac{1}{2}} \\
         \mathbf{\tilde{A}}_k = \mathbf{A}_k + \mathbf{I}\\
         \mathbf{D}_k = diag(\sum_j A_k(u,v))
    \end{array}
\end{equation}

The $l$ consecutive GCN models are connected in a sequential manner through the weights $\mathbf{W}_{k-1}^1$ and $\mathbf{W}_k^1$ of the first convolutional layers \cite{Antaris2020}. For each node $u \in \mathcal{V}_k$ we calculate $h$ independent {\it self-attention heads}, that is vectors $\mathbf{z}_k^j (u) \in \mathbb{R}^{d_1}$, with $j=1,\ldots,h$, based on the $d_1$-dimensional weights $\mathbf{W}_{k-1}^1(u) \in \mathbb{R}^{d_1}$. To compute the weights $\mathbf{W}_k^1$ of each $GCN_k$ model, we average the $h$ independent self-attention vectors $\mathbf{z}_k^j (u)$ \cite{velickovic2018}, as follows:

\begin{equation}
    \begin{array}{l}
    \mathbf{W}_k^1(u) = ELU \big( \frac{1}{h} \sum_{j=1}^{h}{\mathbf{z}_k^j(u)}\big) \\ \\
    \mathbf{z}_k^j(u) =  \sum_{v\in \mathcal{N}_u} \alpha_{u,v} \mathbf{H}_k \mathbf{W}_{k-1}^1(v)
    \end{array}
    \label{eq:self_attention}
\end{equation}
where ELU is the Exponential Linear Unit activation function \cite{elu}. Variable $\mathbf{H}_k \in \mathbb{R}^{d_1 \times d_1}$ is the shared weight transformation matrix applied to the previous weights $\mathbf{W}_{k-1}^1(u)$ of each node $u \in \mathcal{V}_k$, $\mathcal{N}_u$ is the neighborhood set of the node $u$. Variable $\alpha_{u,v}$ is the normalized attention coefficient between $u \in \mathcal{V}_k$ and $v \in \mathcal{N}_u$, which is calculated based on the softmax function \cite{Sankar2020}, as follows:

\begin{equation}
\resizebox{\columnwidth}{!}{
    $\alpha_{u,v} = \frac{exp\big(\sigma(A_k(u,v) \cdot \mathbf{a}_k^T [\mathbf{H}_k \mathbf{W}_{k-1}^1(u) || \mathbf{H}_k \mathbf{W}_{k-1}^1(v)])\big)}{\displaystyle\sum_{w \in \mathcal{N}_u} exp \big( \sigma (A_k(u,w) \cdot \mathbf{a}_k^T [\mathbf{H}_k \mathbf{W}_{k-1}^1(u) || \mathbf{H}_k \mathbf{W}_{k-1}^1(w)]) \big)}$
    }
    \label{eq:att_coef}
\end{equation}

where $\sigma$ is the sigmoid function, $A_k(u,v)$ is the edge weight between $u$ and $v$, $\mathbf{a}_k^T \in \mathbb{R}^{2d_1} $ is a $2d_1$-dimensional weight vector which is applied to the attention process between nodes $u$ and $v$ \cite{Sankar2020, velickovic2018}, and $||$ is the concatenation operation. The attention coefficient $\alpha_{u,v}$ measures the importance of the connection between nodes $u \in \mathcal{V}_k$ and $v \in \mathcal{N}_u$. A high attention coefficient value $\alpha_{u,v}$ corresponds to a connection $e_k(u,v) \in \mathcal{E}_k$ which is maintained over several consecutive graph/viewing snapshots and has high edge weight in the adjacency matrix $A_k(u,v)$. This means that the learned weights $\mathbf{W}_k^1$ reflect on the importance of the existing connection between node $u$ and $v$, processing the convolution accordingly.

To train the teacher model EGAD-$\mathcal{T}$, we initialize $l$ GCN models and connect the consecutive GCN models using the self-attention mechanism in Equation \ref{eq:self_attention}. As aforementioned, each of the $k$-th $GCN$ models takes as an input the normalized adjacency matrix $\tilde{\mathbf{A}}_k$ and the feature vectors $\mathbf{X}_k$. When training EGAD-$\mathcal{T}$, each $GCN_k$ model computes the weights $\mathbf{W}_k^1$ in Equation \ref{eq:self_attention}, and then calculates the latent representations $\mathbf{Z}_k$ based on Equation \ref{eq:gcn}. Note that the weights $\mathbf{W}_0^1$ for the first $GCN$ model are randomly initialized. To train our teacher model EGAD-$\mathcal{T}$, we adopt the Root Mean Square Error loss function with respect to the latent representations $\mathbf{Z}_k$ generated by the last GCN model \cite{Antaris2020}, as follows:

\begin{equation}
    \min_{\mathbf{Z}_k} L^{\mathcal{T}} = \sqrt{\frac{1}{n_k} \big(\sigma({\mathbf{Z}_k^{\mathcal{T}}}^\top \cdot \mathbf{Z}_k^{\mathcal{T}}) - \mathbf{A}_k}\big)^2
    \label{eq:teacher_loss}
\end{equation}
where $\cdot$ represents the inner product operation between all the possible pairs of latent representations, and the term $\sigma({\mathbf{Z}_k^{\mathcal{T}}}^\top \cdot \mathbf{Z}_k^{\mathcal{T}}) - \mathbf{A}_k$ calculates the error of the latent representations $\mathbf{Z}_k^{\mathcal{T}}$ to capture the structure of the graph snapshot $\mathcal{G}_k$. In our implementation, we optimize the parameters $\mathbf{H}_k$ and $\mathbf{a}_k$ between consecutive GCN models, based on the loss function in Equation \ref{eq:teacher_loss} and the backpropagation algorithm.


\subsection{EGAD-$\mathcal{S}$ Student Model} \label{sec:egad_student}

We train the student model EGAD-$\mathcal{S}$ to compute the online latent representations $\mathbf{Z}_k^{\mathcal{S}}$, by exploiting the knowledge of the pretrained teacher model EGAD-$\mathcal{T}$. As we train the student model only on online data, the student model EGAD-$\mathcal{S}$ requires significantly less number of trainable parameter weights, compared with the teacher model EGAD-$\mathcal{T}$. The student model EGAD-$\mathcal{S}$ consists of $l$ consecutive GCN models, with EGAD-$\mathcal{S} = \{GCN_{k-l},\ldots,GCN_k\}$. We calculate the weights $\mathbf{W}_k^1$ (Equation \ref{eq:att_coef}), and compute the latent representations based on Equation \ref{eq:gcn}. 

The knowledge acquired by the teacher model EGAD-$\mathcal{T}$ is transferred to the student model EGAD-$\mathcal{S}$ via the distillation loss function $L^{\mathcal{D}}$, adopted by the student model during the online training process. We formulate the distillation loss function as a minimization problem for the student model EGAD-$\mathcal{S}$ as follows:

\begin{equation}
    \min_{\mathbf{Z}_k^{\mathcal{S}}} L^{\mathcal{D}} = (1 - \gamma) L^{\mathcal{T}} + \gamma L^{\mathcal{S}} 
    \label{eq:student_loss}
\end{equation}
where $L^{\mathcal{T}}$ is the inference error of the teacher model in Equation \ref{eq:teacher_loss}, and $L^{\mathcal{S}}$ is the root mean squared error with the latent representations $\mathbf{Z}_k^{\mathcal{S}}$ generated by the student model. Hyper-parameter $\gamma \in [0,1]$ balances the training of the student model EGAD-$\mathcal{S}$ when inferring the knowledge of the teacher model EGAD-$\mathcal{T}$. A higher value of $\gamma$ emphasizes more on the student model EGAD-$\mathcal{S}$ and distillates less knowledge from the teacher model EGAD-$\mathcal{T}$. The distillation loss function $L^{\mathcal{D}}$ in Equation \ref{eq:student_loss} allows the student model EGAD-$\mathcal{S}$ to overcome any bias introduced by the teacher model EGAD-$\mathcal{T}$ \cite{hinton2015, Bucilua2006, Jiaxi2018, Mary2019}. This means that EGAD-$\mathcal{S}$ can achieve similar or better accuracy than the teacher model EGAD-$\mathcal{T}$. As we will show later in Section \ref{sec:distil}, the student model EGAD-$\mathcal{S}$ consistently outperforms the teacher model EGAD-$\mathcal{T}$ in terms of accuracy, by significantly downsizing the number of parameters. 

\section{Experimental Evaluation} \label{sec:exp_eval}

\subsection{Evaluation Setup}

In our experiments we evaluate the performance of the proposed model on the link prediction task. To examine the two different components of our model, we train the teacher and student models EGAD-$\mathcal{T}$ and EGAD-$\mathcal{S}$, using $l$ consecutive graph/viewing snapshots up to the $k$-th graph $\mathcal{G}_{k}$. The task of link prediction is to forecast the unobserved connections, denoted by $\mathcal{O}_{k+1} = \mathcal{E}_{k+1} \backslash \{\mathcal{E}_{k-l},\ldots,\mathcal{E}_k\}$, that will occur in the next graph/viewing snapshot $\mathcal{G}_{k+1}$. Following the evaluation protocol of \cite{Sankar2020, Pareja2020, mahdavi2019}, we concatenate the latent representations $\mathbf{Z}_k(u)$ and $\mathbf{Z}_k(v)$ based on the Hadamard operator, for the unobserved connection $o(u,v) \in \mathcal{O}_{k+1}$ of the viewers $u$ and $v\in\mathcal{V}_k$. The concatenated latent representations are then applied to a Multi-Layer Perceptron (MLP), to calculate the weight of the connection. To measure the online inference efficiency, we report the number of parameters that each model requires to train. Moreover, regarding the prediction accuracy we evaluate the examined models based on the metrics Root Mean Squared Error (RMSE) and Mean Absolute Error (MAE):

\begin{equation}
    \begin{array}{l}
    MAE = \frac{\displaystyle\sum_{o(u,v) \in \mathcal{O}_{k+1}} \left|A_{k+1} (u,v) - \mathbf{Z}_k^T(u) \mathbf{Z}_k(v) \right|}{|\mathcal{O}_{k+1}|} \\ \\
    \resizebox{\columnwidth}{!}{
    $RMSE = \sqrt{(\frac{1}{|\mathcal{O}_{k+1}|})\displaystyle\sum_{o(u,v) \in \mathcal{O}_{k+1}} {(\mathbf{Z}^T_k(u)  \mathbf{Z}_k(v)  - \mathbf{A}_{k+1}(u,v))^{2}}}$
    }
    \end{array}
\end{equation}
Following \cite{Sankar2020, Pareja2020, Antaris2020}, for each snapshot $k$ we train each examined model on $l$ previous graph/viewing snapshots $\mathcal{G}_{k-l}, \ldots, \mathcal{G}_k$, which are considered the offline data for each time step. We randomly select $20\%$ of the unobserved links $\mathcal{O}_{k+1}$ for validation set to tune the model hyper-parameters. The remaining $80\%$ of the unobserved links are considered as the test set, which are the online data for each time step. We repeated our experiments five times, and we report the average RMSE and MAE over the five trials.

\subsection{Examined Models} \label{sec:exam_model}
We compare the performance of the proposed EGAD-$\mathcal{T}$ and EGAD-$\mathcal{S}$ models with the following baseline strategies:
\begin{itemize} 
    \item \textbf{DynVGAE}~\cite{mahdavi2019} is a dynamic joint learning model that shares the trainable parameters between consecutive variational graph auto-encoders \cite{kipf2017}. We implemented DynVGAE from scratch and publish our code\footnote{\url{https://github.com/stefanosantaris/DynVGAE}}, as there is no publicly available implementation. 
    
    \item \textbf{EvolveGCN}\footnote{\url{https://github.com/IBM/EvolveGCN}}~\cite{Pareja2020} is a dynamic graph representation learning model with Gated Recurrent Units (GRUs) between the convolutional weights of consecutive GCNs. 
    
    \item \textbf{DySAT}\footnote{\url{https://github.com/aravindsankar28/DySAT}}~\cite{Sankar2020} is a dynamic self-attention model that captures the evolution of the graph using multi-head self-attention between consecutive graph snapshots. 
    
    \item \textbf{DMTKG-$\mathcal{T}$}~\cite{Jiaqi2019} is the teacher model of the DMTKG knowledge distillation strategy. DMTKG-$\mathcal{T}$ employs Heat Kernel Signature (HKS) on static graph/viewing snapshots and uses Convolutional Neural Network layers to calculate the latent representations based on DeepGraph \cite{li2016deepgraph}. To ensure fair comparison, we train DMTKG-$\mathcal{T}$ per snapshot, with each snapshot containing aggregated graph history up to $k$-th snapshot. As the source code of the DMKTG distillation strategy is not available, we made our implementation publicly available\footnote{\url{https://github.com/stefanosantaris/DMTKG}}.
    
    \item \textbf{DMTKG-$\mathcal{S}$}~\cite{Jiaqi2019} is the student model of the DMTKG strategy, where the goal is to minimize a distillation loss function based on the weighted cross entropy. 
\end{itemize}

\begin{figure*}[t] \centering
\begin{tabular}{ccc}
\includegraphics[scale=0.2]{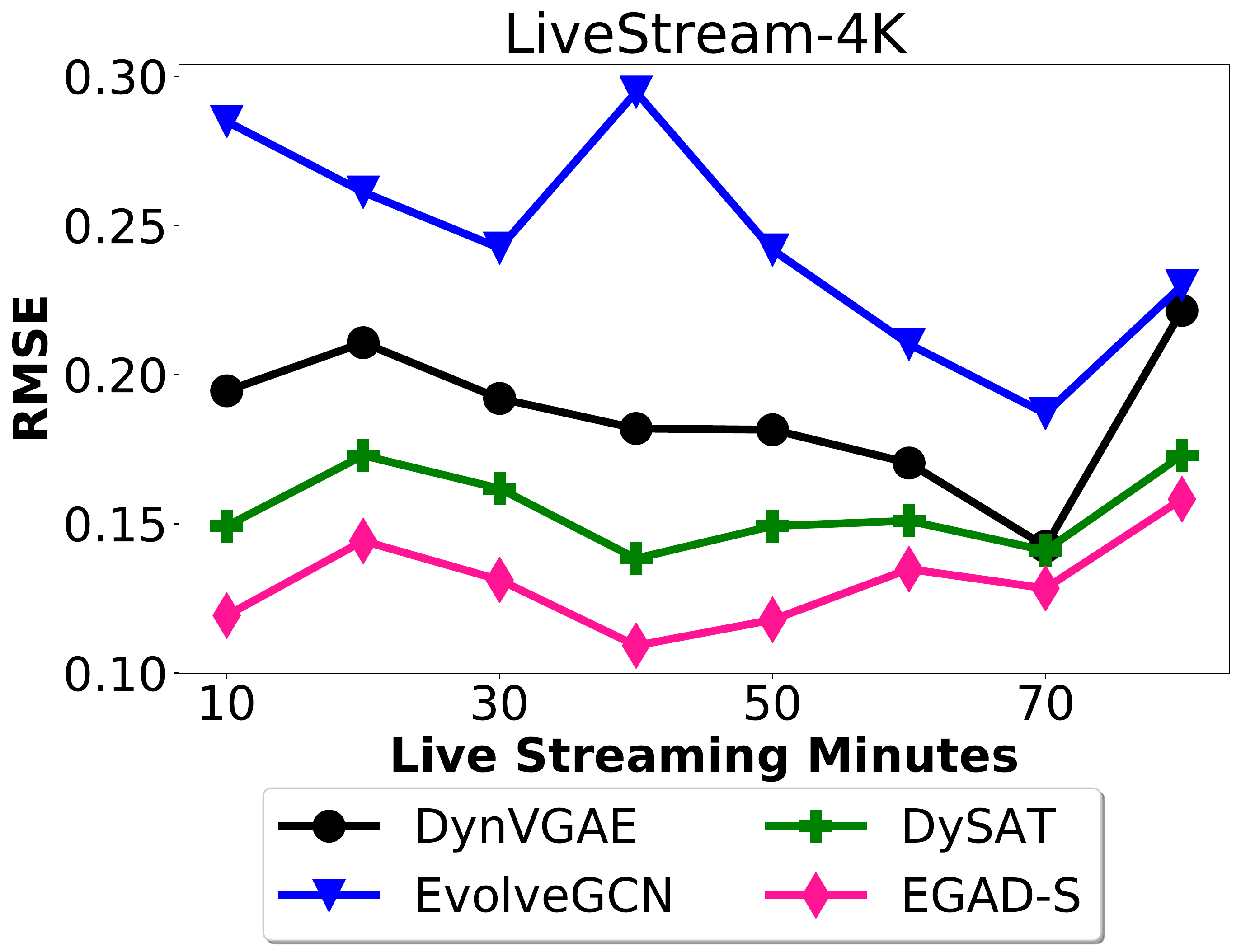} &
\includegraphics[scale=0.2]{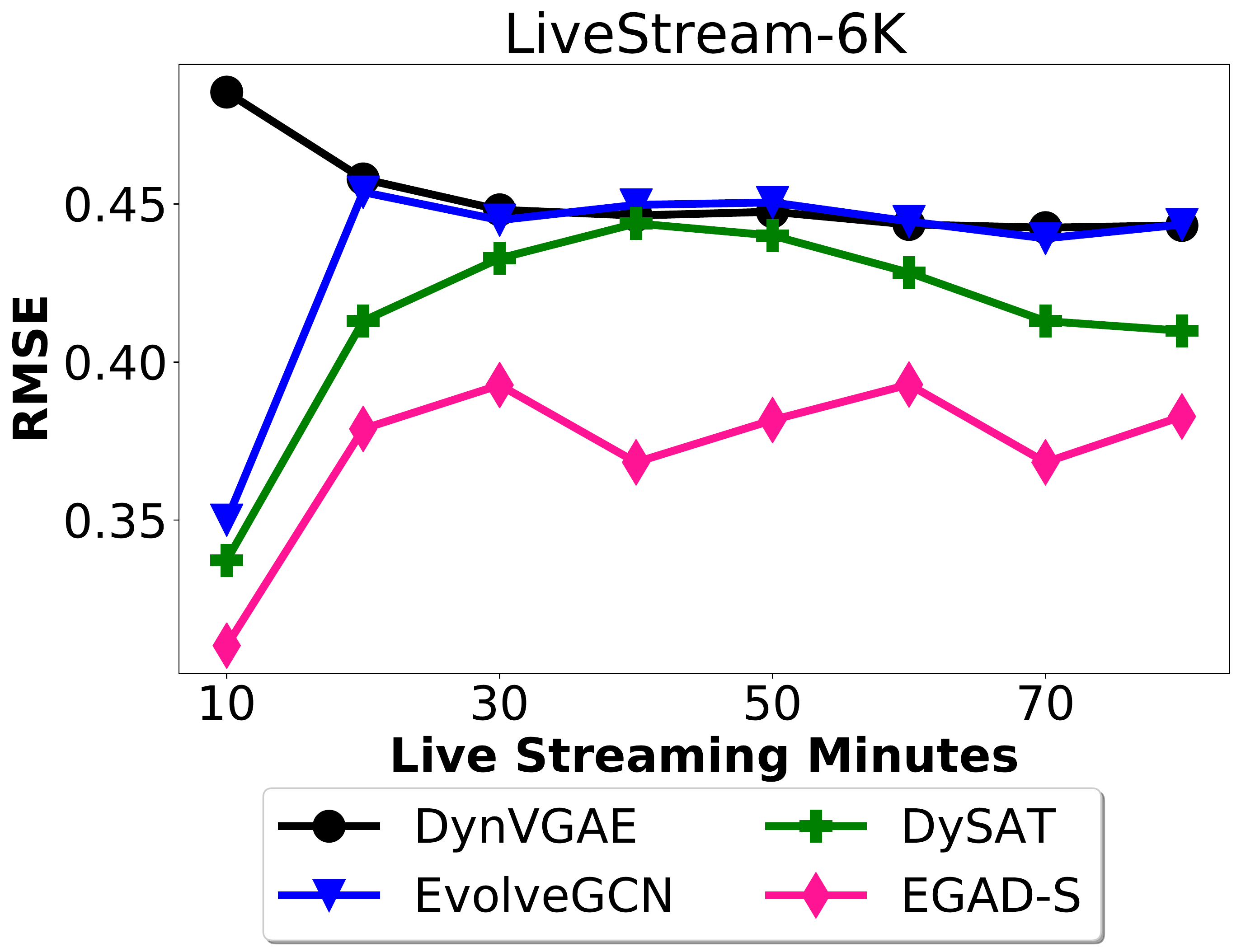} & 
\includegraphics[scale=0.2]{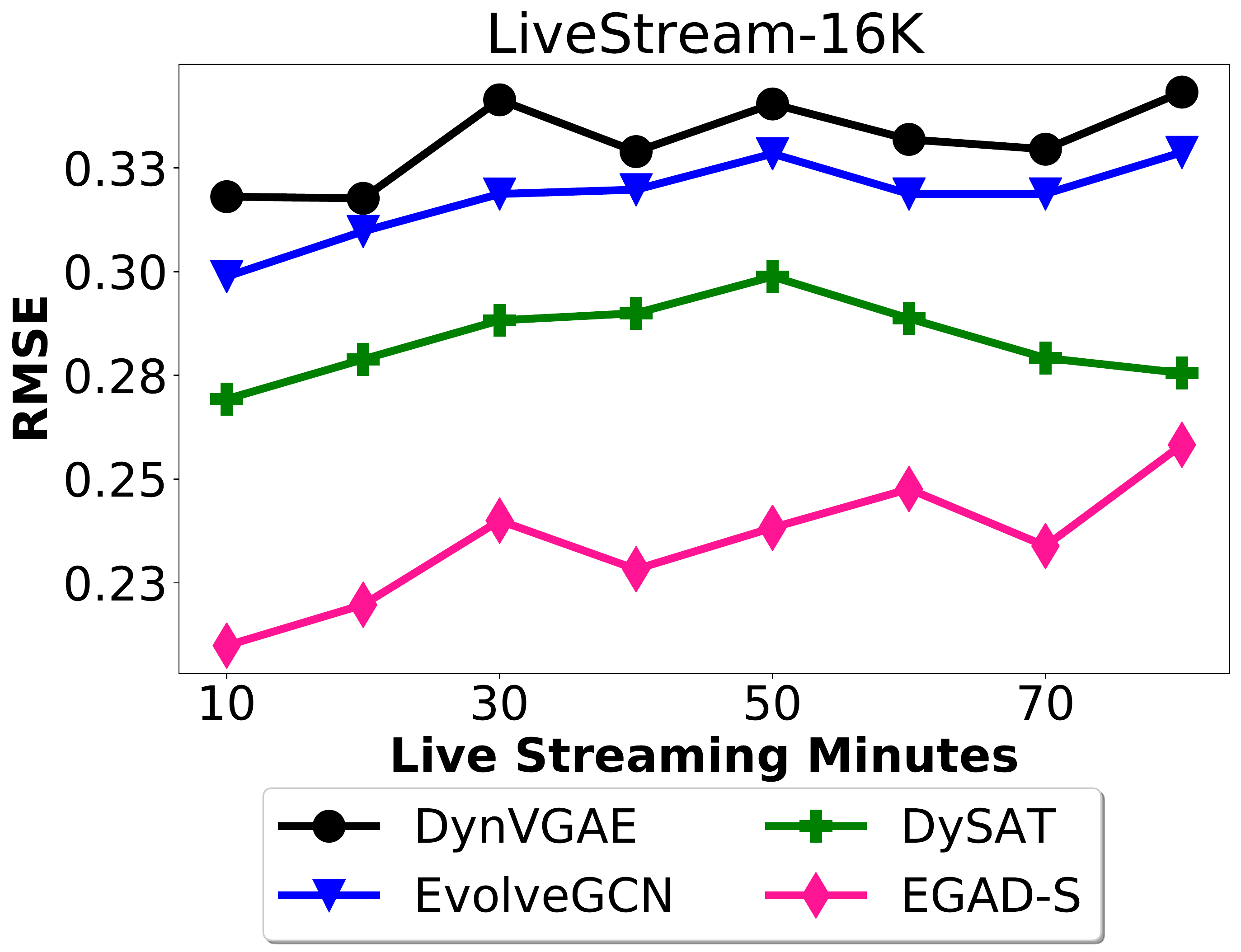} \\

\includegraphics[scale=0.2]{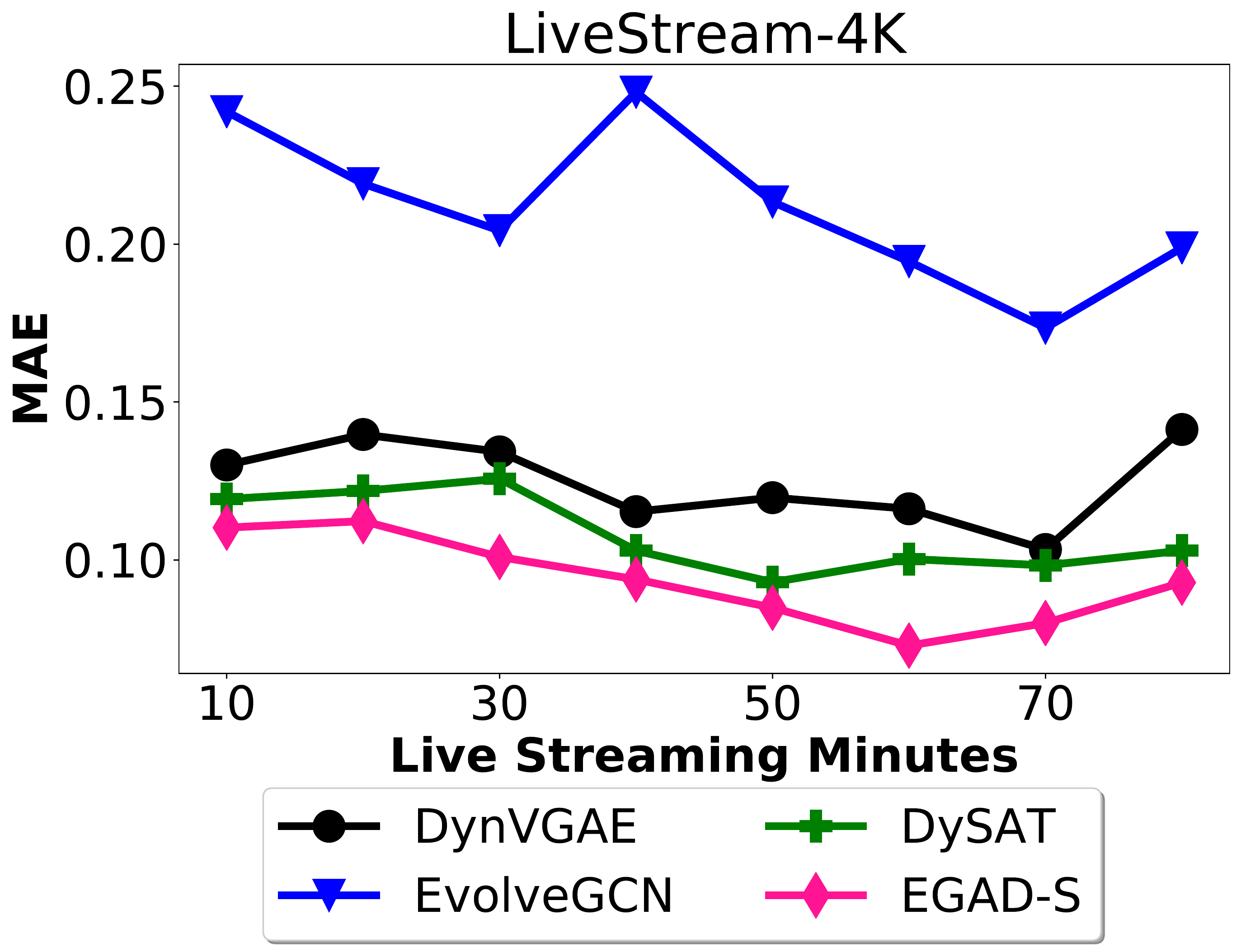} &
\includegraphics[scale=0.2]{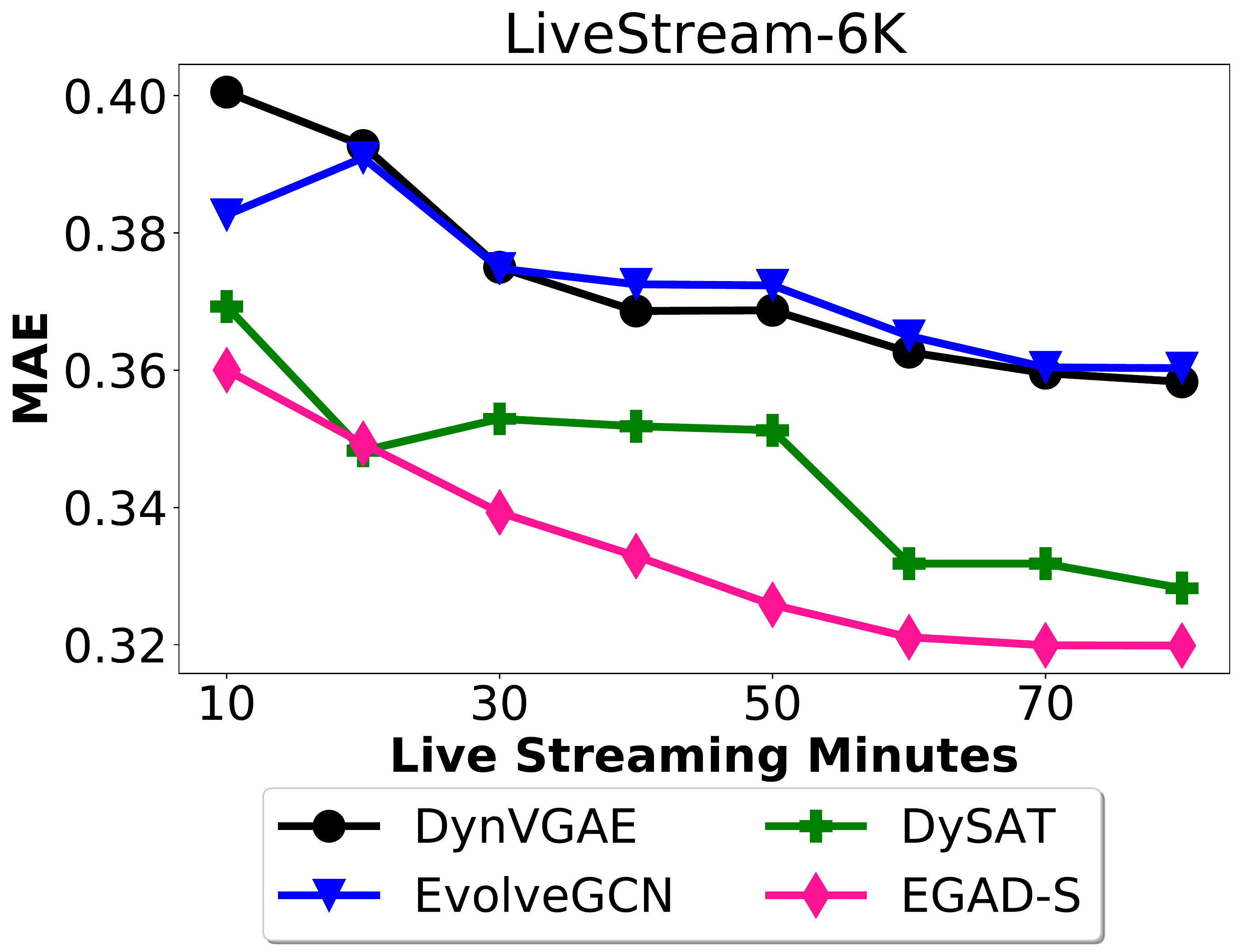} & 
\includegraphics[scale=0.2]{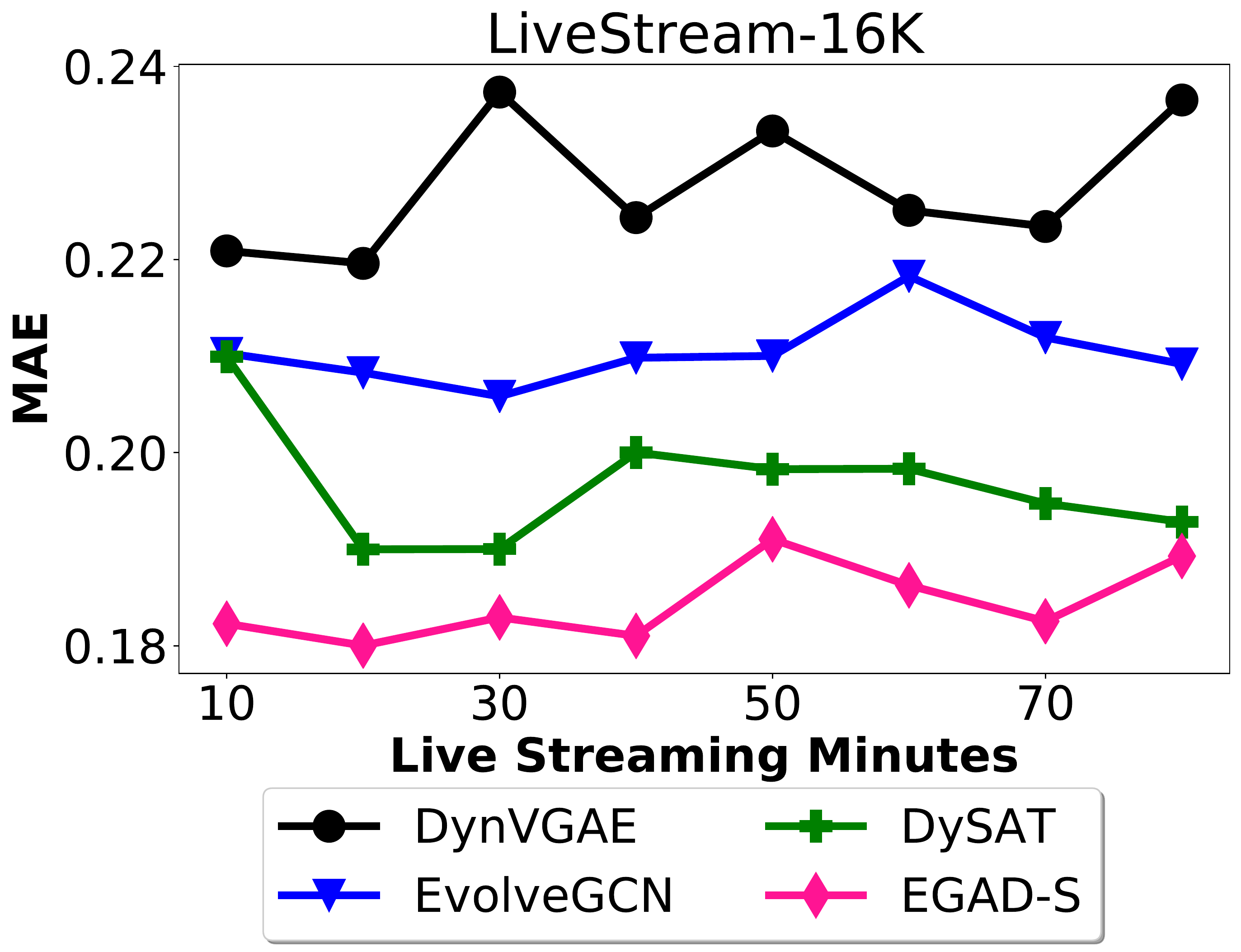} \\

\end{tabular}

\vspace{-0.2cm}
\caption{Performance evaluation of EGAD-$\mathcal{S}$ against the non-distillation strategies in terms of RMSE and MAE in LiveStream-4K, LiveStream-6K and LiveStream-16K.} \label{fig:rmse_mae}
\end{figure*}

\begin{table*}[ht]
    \caption{Model parameters in millions for the live streaming minutes, when comparing the proposed EGAD-$\mathcal{S}$ model with the non-distillation strategies.}
    
    \vspace{-0.2cm}
    
\resizebox{\textwidth}{!}{
    \centering
    \begin{tabular}{|c|cccc|cccc|cccc|}
    \hline
         & \multicolumn{4}{c|}{\textbf{LiveStream-4K}} & \multicolumn{4}{c|}{\textbf{LiveStream-6K}} & \multicolumn{4}{c|}{\textbf{LiveStream-16K}} \\ \hline\hline
         \textbf{Stream. Min.} & \textbf{DynVGAE} & \textbf{EvolveGCN} & \textbf{DySAT} & \textbf{EGAD-$\mathcal{S}$} & \textbf{DynVGAE} & \textbf{EvolveGCN} & \textbf{DySAT} & \textbf{EGAD-$\mathcal{S}$} & \textbf{DynVGAE} & \textbf{EvolveGCN} & \textbf{DySAT} & \textbf{EGAD-$\mathcal{S}$} \\\hline
         10 & 0.332 & 37.034  & 0.845 & \textbf{0.081} & 0.092 & 2.288   & 0.367 & \textbf{0.022} & 0.152 & 6.994    & 0.485 & \textbf{0.036} \\\hline
         20 & 0.830 & 117.829 & 1.227 & \textbf{0.105} & 0.902 & 70.021  & 1.299 & \textbf{0.114} & 0.626 & 65.472   & 1.023 & \textbf{0.079} \\\hline
         30 & 0.903 & 140.150 & 1.515 & \textbf{0.117} & 1.312 & 151.137 & 1.924 & \textbf{0.168} & 1.220 & 260.283  & 1.832 & \textbf{0.156} \\\hline
         40 & 0.974 & 154.609 & 1.774 & \textbf{0.125} & 1.458 & 187.345 & 2.285 & \textbf{0.189} & 1.907 & 646.523  & 2.734 & \textbf{0.245} \\\hline
         50 & 0.973 & 163.519 & 1.800 & \textbf{0.128} & 1.567 & 216.853 & 2.394 & \textbf{0.202} & 2.678 & 1285.467 & 3.510 & \textbf{0.341} \\\hline
         60 & 0.988 & 168.607 & 1.815 & \textbf{0.130} & 1.641 & 238.277 & 2.468 & \textbf{0.212} & 3.503 & 2211.020 & 4.330 & \textbf{0.445} \\\hline
         70 & 0.100 & 172.770 & 1.827 & \textbf{0.132} & 1.670 & 255.987 & 2.527 & \textbf{0.219} & 3.986 & 2867.735 & 3.813 & \textbf{0.505} \\\hline
         80 & 1.001 & 176.247 & 1.836 & \textbf{0.133} & 1.737 & 267.638 & 2.564 & \textbf{0.224} & 4.392 & 3486.341 & 5.219 & \textbf{0.556} \\\hline
        
    \end{tabular}
    }

    \label{tab:parameters}
\end{table*}

\noindent
\textbf{Settings.} In Tables III-V we report the performance of each examined model in terms of RMSE when calibrating the hyper-parameters of the examined models, following a cross-validation strategy. For each model, we tuned the hyper-parameters based on a grid selection strategy and select the best configuration. In particular, in DynVGAE we set the representation size to $d = 64$ and the window size $l = 2$ for all datasets. In EvolveGCN, the representation size is set to $d=32$, with $l=2$ previous graph/viewing snapshots. DySAT uses $l=2$ consecutive graph/viewing snapshots and employs $h=3$ attention heads for the LiveStream-4K and LiveStream-6K datasets. For the LiveStream-16K dataset, we use $h=4$ attention heads. The representation size is fixed to $d=64$ in all datasets. In DMTKG-$\mathcal{T}$, the representation size is fixed to $d=64$ for LiveStream-4K and LiveStream-6K, while in LiveStream-16K we use $d=128$. In DMTKG-$\mathcal{S}$, we reduce the model size by setting the size of latent representations to $d=32$ for LiveStream-4K and LiveStream-6K, and $d=64$ for LiveStream-16K. Regarding the proposed model, we train EGAD-$\mathcal{T}$  on $l=3$ consecutive graph/viewing snapshots and set the number of head attentions to $h=3$ in Equation \ref{eq:self_attention}, with $d=64$-dimensional latent representations. In the student model EGAD-$\mathcal{S}$, we reduce the number of heads $h=1$ and fix the latent representation size to $d=16$ for all datasets. The influence of $\gamma$ on the distillation loss function (Equation~\ref{eq:student_loss}) is further studied in Section~\ref{sec:distil}. We initialize the learning rate to $1e-03$ based on the Adam Optimizer with $200$ epochs. All experiments were performed on an Intel(R) Xeon(R) Bronze 3106 CPU 1.70GHz machine and GPU accelerated with the GEFORCE RTX 2080 Ti graph card.

\subsection{Performance Evaluation} \label{sec:perf_eval}

In Figure \ref{fig:rmse_mae}, we evaluate the performance of the student model EGAD-$\mathcal{S}$ against the non-distillation strategies, that is DynVGAE, EvolveGCN and DySAT, in terms of RMSE and MAE. We observe that all models have a higher prediction error in terms of RMSE and MAE in LiveStream-6K than the other datasets. This occurs because the viewers in the LiveStream-6K dataset attended the live video streaming event in a completely different pattern (Section \ref{sec:data_analysis}) than LiveStream-4K and LiveStream-16K. More precisely, in LiveStream-6K the number of viewers that emerge in $0$-$20$ minutes is significantly higher than the other events, which negatively impacts the prediction accuracy of the examined models. 


The student model EGAD-$\mathcal{S}$ significantly outperforms the baseline approaches in all datasets. This suggests that the proposed student model EGAD-$\mathcal{S}$ can efficiently capture the evolution of the graph in the learned latent representations $\mathbf{Z}_k^{\mathcal{S}}$. The second best approach is DySAT, demonstrating the ability of self-attention mechanisms to generate accurate latent representations. DySAT calculates the latent representations $\mathbf{Z}_k$ by applying self-attentional aggregations to the local node neighborhoods. Instead, the proposed EGAD-$\mathcal{S}$ model performs self-attention to the convolutional weights between consecutive GCNs. Thus, our model is able to efficiently capture the different graph evolution patterns of the live video streaming events. Compared to the second best method DySAT, the proposed EGAD-$\mathcal{S}$ model achieves relative drops $9.8$ and $13.5\%$ in terms of RMSE and MAE in the LiveStream-4K dataset. Similarly, EGAD-$\mathcal{S}$ achieves relative drops $10.2$ and $3.5\%$ in LiveStream-6K, and $17.3$ and $6.2\%$ relative drops in LiveStream-16K.

\begin{figure*}[ht] \centering
\begin{tabular}{ccc}
\includegraphics[scale=0.2]{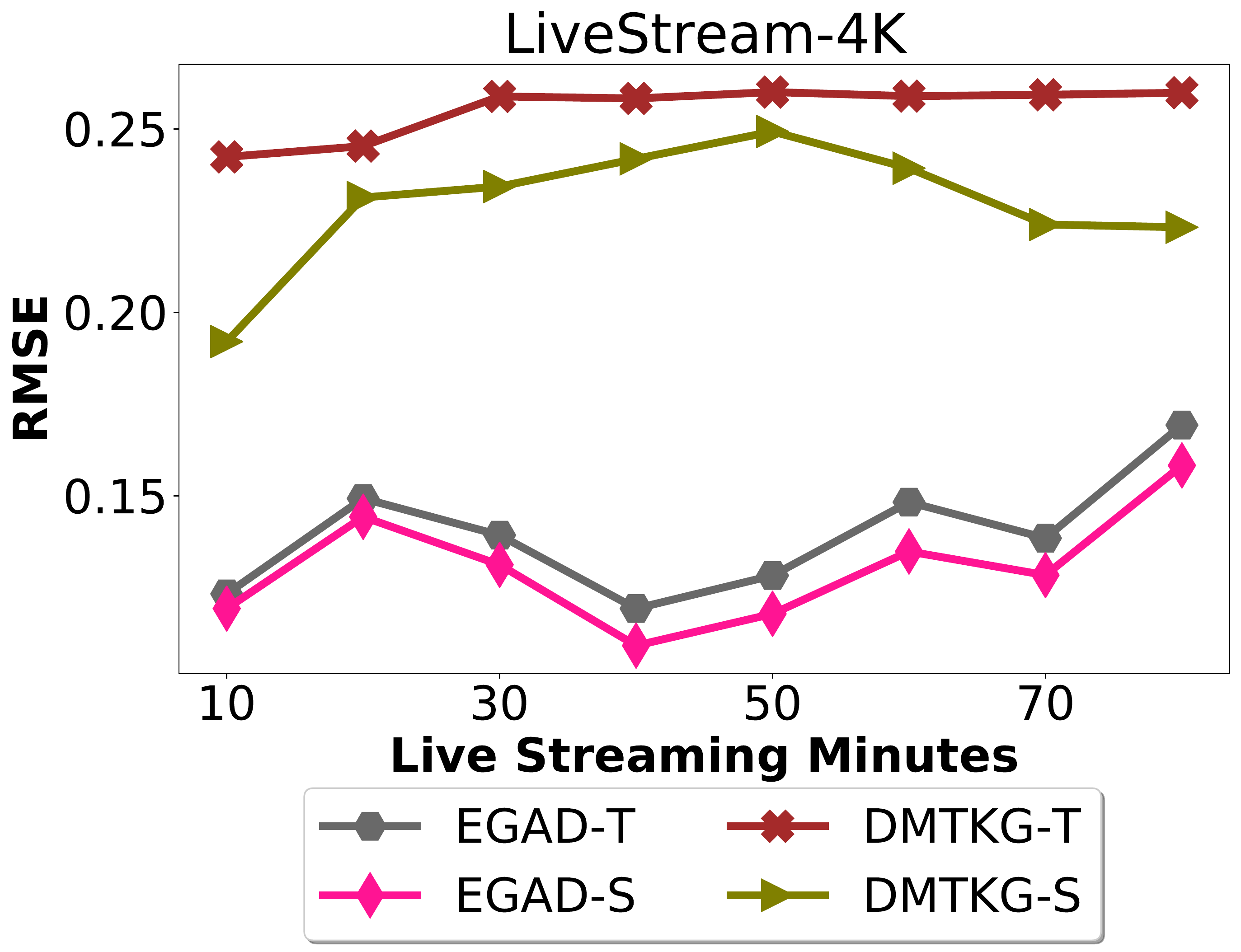} &
\includegraphics[scale=0.2]{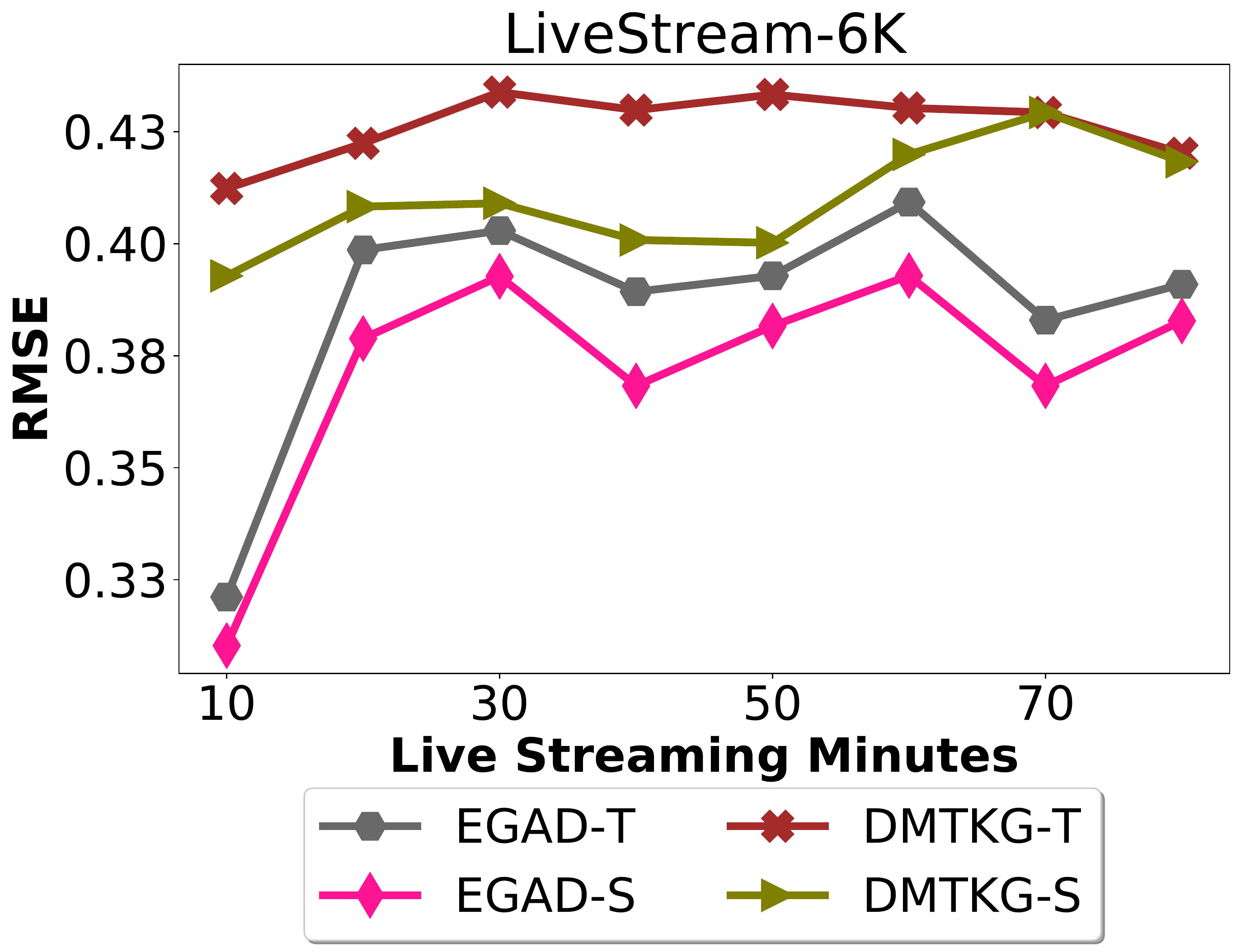} & 
\includegraphics[scale=0.2]{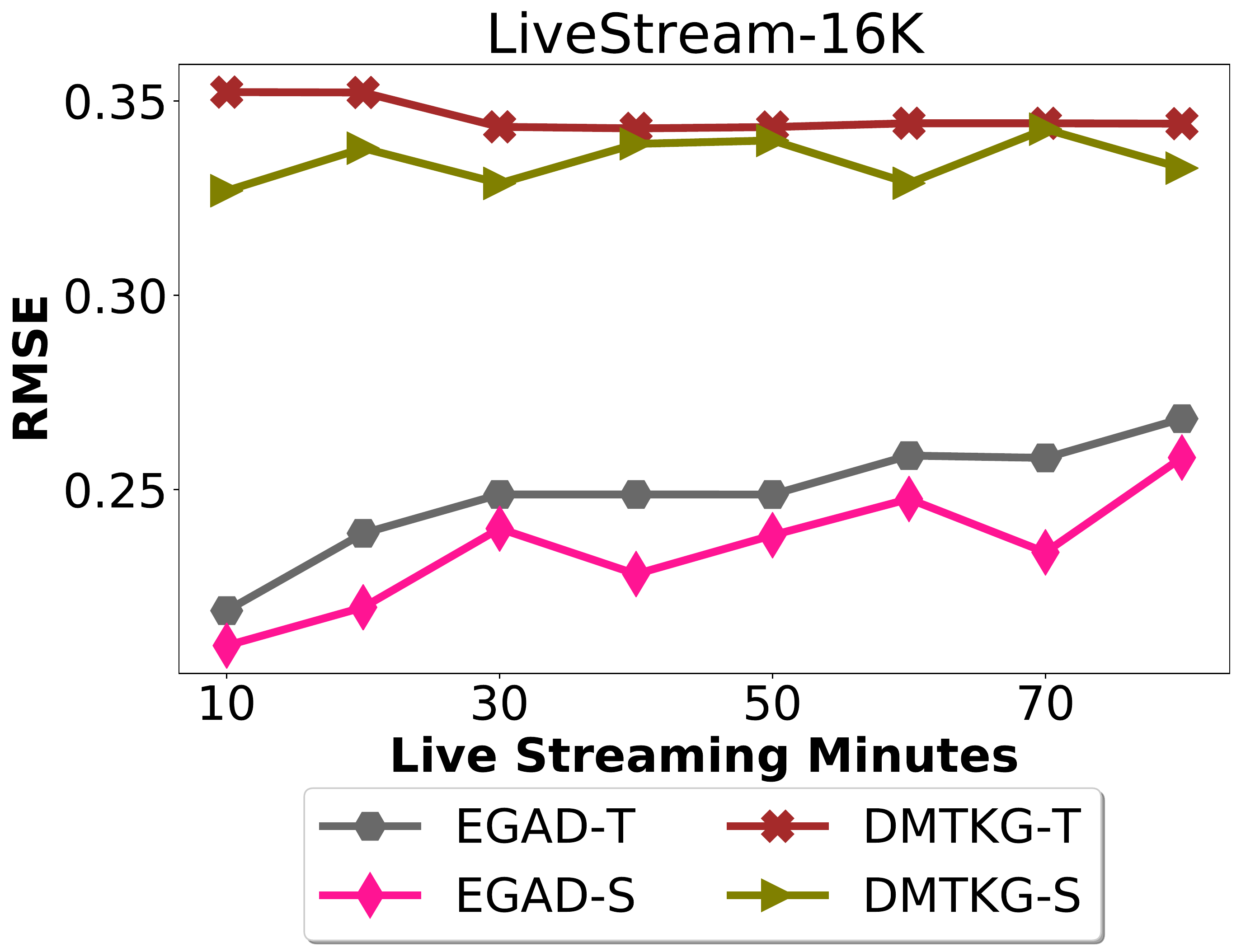}
\end{tabular}

\vspace{-0.2cm}
\caption{Comparison of student and teacher models for the examined knowledge distillation strategies.} \label{fig:teacher_student_rmse}
\end{figure*}

\begin{figure*}[ht] \centering
\begin{tabular}{ccc}
\includegraphics[scale=0.2]{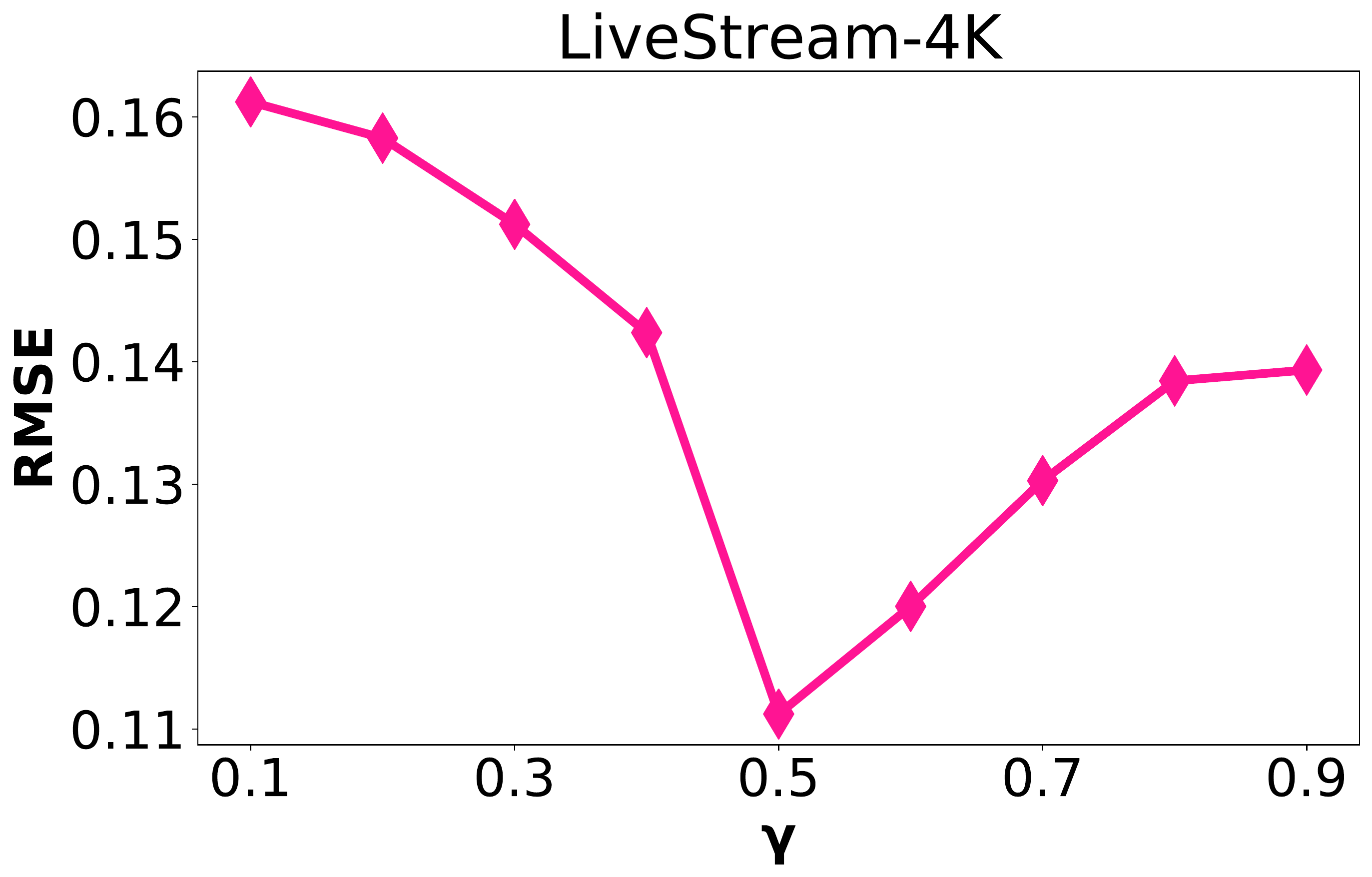} &
\includegraphics[scale=0.2]{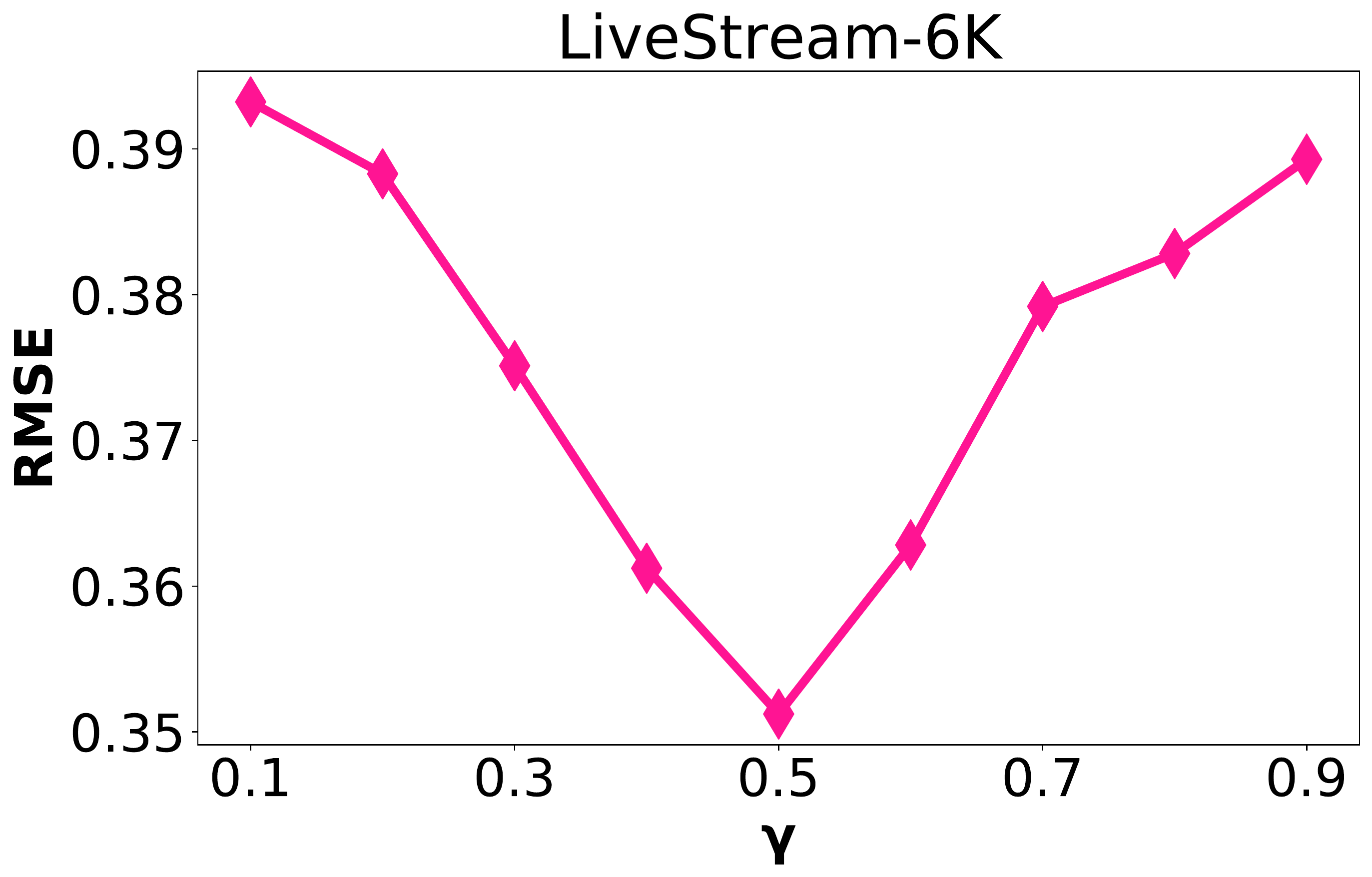} & 
\includegraphics[scale=0.2]{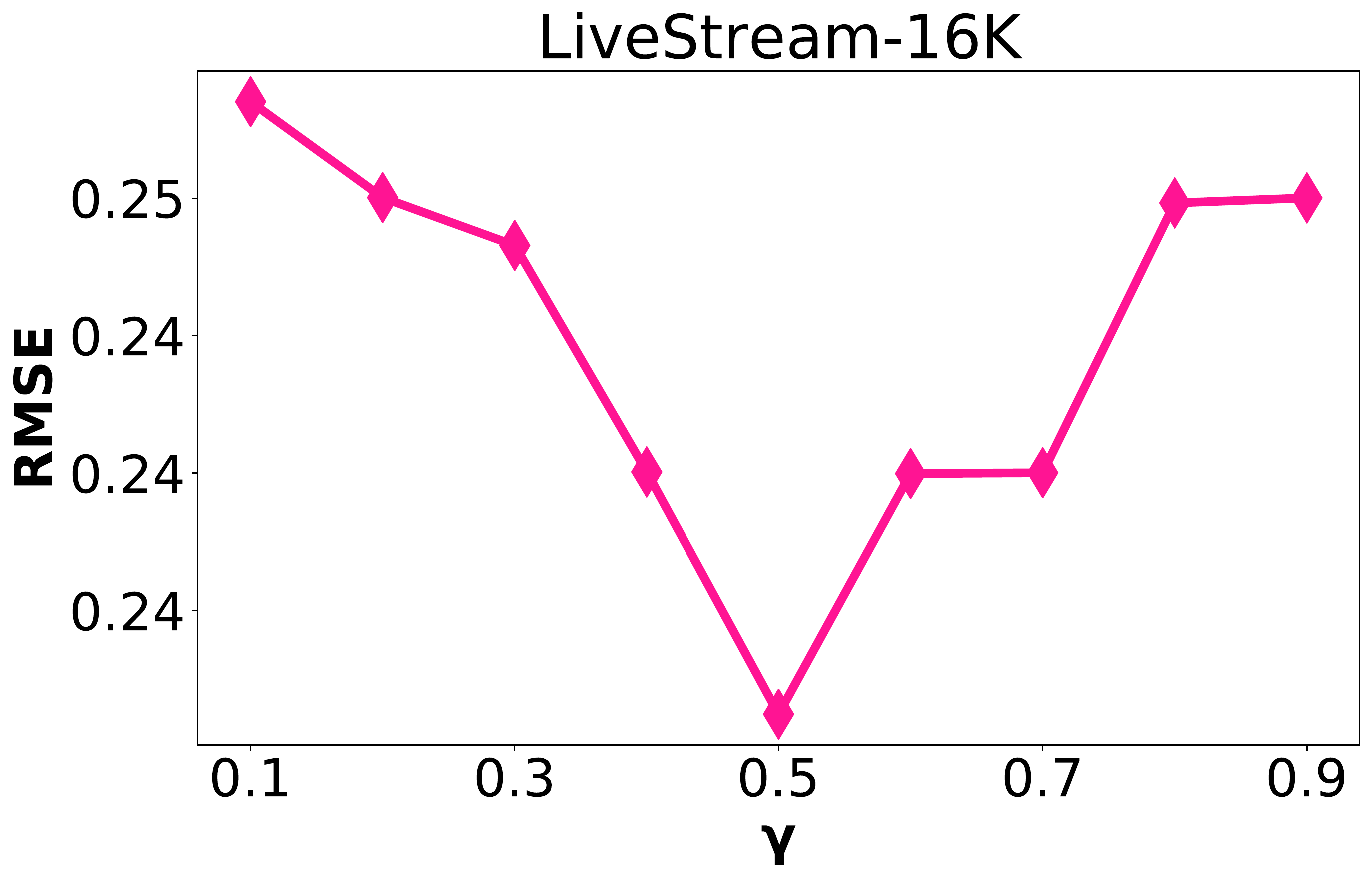}
\end{tabular}

\vspace{-0.2cm}
\caption{Impact of $\gamma$ on the prediction accuracy of student model EGAD-$\mathcal{S}$.} \label{fig:gamma}
\end{figure*}

In Table \ref{tab:parameters}, we present the numbers of parameters in millions that are required to train the examined models. As aforementioned in Section~\ref{sec:data_analysis}, the majority of the viewers in the LiveStream-4K dataset started to attend the live video streaming event from the first $10$ minutes. Therefore, all models have fewer trainable parameters on the first graph snapshots $k=0-30$ minutes in LiveStream-6K and LiveStream-16K than in the LiveStream-4K dataset. We observe that EGAD-$\mathcal{S}$ clearly outperforms the baseline approaches in terms of the required parameters. Evaluated against DynVGAE, EvolveGCN and DySAT, the average compress ratios of the student model EGAD-$\mathcal{S}$ are $12$:$100$, $7$:$1000$, and $7$:$100$, respectively. Provided that EGAD-$\mathcal{S}$ constantly outperforms all the baseline approaches in terms of RMSE and MAE, the high compression ratios demonstrate the ability of the proposed knowledge distillation strategy to significantly reduce the model size in terms of required parameters. Moreover, it is clear that EvolveGCN model requires a significant amount of trainable parameters to generate the latent representations. This means that EvolveGCN does not scale well when increasing the number of viewers in live video streaming events. As DySAT employs multi-head attention on consecutive graph/viewing snapshots, and not on consecutive GCNs as the proposed EGAD- model does, DySAT requires a much a higher number of parameters by following a non-distillation strategy.

\subsection{Distillation Evaluation} \label{sec:distil}

In Figure \ref{fig:teacher_student_rmse}, we study the impact of the proposed knowledge distillation strategy on the student model EGAD-$\mathcal{S}$ in terms of RMSE, when compared with the teacher model EGAD-$\mathcal{T}$. In addition, in this set of experiments we evaluate our model against DMTKG \cite{Jiaqi2019}, a baseline graph representation approach with knowledge distillation, comparing with both the teacher model DMTKG-$\mathcal{T}$ and student model DMTKG-$\mathcal{S}$.

On inspection of Figure \ref{fig:teacher_student_rmse}, we observe that the EGAD-$\mathcal{T}$ and EGAD-$\mathcal{S}$ models outperform DMTKG-$\mathcal{T}$ and DMTKG-$\mathcal{S}$ in all datasets. This occurs because DMTKG applies knowledge distillation on top of DeepGraph \cite{li2016deepgraph}, which is a static graph representation learning approach. Therefore, DMTKG ignores the graphs' evolution when learning the latent representations. An interesting observation is that the student models EGAD-$\mathcal{S}$ and DMTKG-$\mathcal{S}$ achieve higher performance than the respective teacher models EGAD-$\mathcal{T}$ and DMTKG-$\mathcal{T}$. This indicates the effectiveness of the examined distillation strategies to correctly transfer the knowledge of the teacher models to the respective student models. This occurs because the student models remove the bias of the teacher models to the offline data, and achieve high prediction accuracy, complying with similar observations that have been made in relevant studies \cite{Jiaxi2018,kim2016sequencelevel}. Compared to the EGAD-$\mathcal{T}$ model, EGAD-$\mathcal{S}$ achieves $6.5$, $3.6$ and $5.7\%$ relative drops in terms of RMSE for LiveStream-4K, LiveStream-6K and LiveStream-16K, respectively. 

In Table \ref{tab:distil_parameters}, we present the maximum number of parameters in millions that are required to train the examined models during the live video streaming events. EGAD-$\mathcal{S}$ significantly reduces the number of required parameters, achieving compression ratios $15$:$100$, $17$:$100$ and $21$:$100$, on average, in LiveStream-4K, LiveStream-6K and LiveStream-16K, respectively. This occurs because the student model EGAD-$\mathcal{S}$ uses a lower number of attention heads $h$ and representation size $d$ than the teacher model EGAD-$\mathcal{T}$ (Section \ref{sec:perf_eval}). Therefore, EGAD-$\mathcal{S}$ has lower online inference latency, compared with the teacher model EGAD-$\mathcal{T}$. Instead, the DMTKG distillation strategy achieves an average $1$:$2$ compression ratio for the student model DMTKG-$\mathcal{S}$. The DMTKG distillation strategy is not able to further reduce the student model size, because DMTKG is designed for static graphs. This indicates that the DMTKG-$\mathcal{S}$ model requires more trainable parameters to learn accurate latent representations than the proposed EGAD-$\mathcal{S}$ model.

\begin{table}[h]
    \caption{The maximum numbers of required parameters in millions of the examined knowledge distillation strategies during the live video streaming events. In the parentheses, we denote the average compression ratios of the student models, when compared with the respective teacher models.}
    
    \vspace{-0.2cm}
    
    \resizebox{0.46\textwidth}{!}{
        \centering
        \begin{tabular}{|c|c|c|c|}
        \hline
            \textbf{Model} & \textbf{LiveStream-4K} & \textbf{LiveStream-6K} & \textbf{LiveStream-16K} \\\hline
            \textbf{DMTKG}-$\mathcal{T}$ & 3.673 & 5.129 & 10.437 \\\hline
            \textbf{DMTKG}-$\mathcal{S}$ & 1.836 (1:2) & 2.564 (1:2) & 5.219 (1:2) \\\hline
            \textbf{EGAD}-$\mathcal{T}$ & 0.918 & 1.282 & 2.609\\\hline
            \textbf{EGAD}-$\mathcal{S}$ & 0.133 (15:100) & 0.224 (17:100) & 0.556 (21:100) \\\hline
        \end{tabular}
    }

    \label{tab:distil_parameters}
\end{table}

In Figure \ref{fig:gamma}, we evaluate the influence on the hyper-parameter $\gamma$ of Equation \ref{eq:student_loss} on the student model EGAD-$\mathcal{S}$. We vary the hyper-parameter $\gamma$ from $0.1$ to $0.9$ by a step of $0.1$, to balance the impact of the student $L^{\mathcal{S}}$ and teacher $L^{\mathcal{T}}$ losses on the distillation loss function $L^{\mathcal{D}}$. For each parameter $\gamma$, we report the averaged RMSE over all the graph snapshots of the live video streaming events. In all datasets, the student model EGAD-$\mathcal{S}$ achieves the highest performance when we equally balance the influence of the student and teacher models ($\gamma=0.5$). For larger values of parameter $\gamma$, the student model EGAD-$\mathcal{S}$ emphasizes more on the loss $L^{\mathcal{S}}$ than the loss $L^{\mathcal{T}}$. As a consequence, the student model EGAD-$\mathcal{S}$ distills less knowledge from the teacher model EGAD-$\mathcal{T}$, which negatively impacts the performance of the EGAD-$\mathcal{S}$ model in terms of RMSE. Instead, decreasing the hyper-parameter $\gamma$ prevents the student model EGAD-$\mathcal{S}$ from training on the online graph data and at the same time introduces the bias to the offline data of EGAD-$\mathcal{T}$. This means that for small values of $\gamma$ EGAD-$\mathcal{S}$ mainly distills the knowledge of the teacher model EGAD-$\mathcal{T}$, resulting in limited prediction accuracy.

\section{Conclusion} \label{sec:conc}

In this paper, we presented a knowledge distillation strategy, to overcome the problem of online latency inference of dynamic graph representation learning approaches in live video streaming events. Evaluated against several baseline approaches on three real-world live video streaming events, the proposed model achieves $7$:$100$ compression ratio on average. Moreover, the proposed student model preserves high prediction accuracy, achieving average relative drops $12.4$ and $7.7\%$ in terms of RMSE and MAE in all events, when compared with the second best approach. Distributed live video streaming providers, such as Hive Streaming AB, can significantly benefit from our model by significantly reducing the required parameters/computational time in the link prediction task. In doing so, viewers can exploit the offices' internal high bandwidth network from the beginning of the live video streaming event, by avoiding to establish low bandwidth connections. Provided that several offices have limited network capacity, our model can significantly reduce the generated network traffic. Therefore, enterprises can distribute high quality video content to their offices without any network limitations, improving user experience. Moreover, the proposed model allows enterprises to distribute video content of high resolution, such as 4K. 


There are several interesting future directions to graph representation learning for live video streaming events.
\begin{itemize}
    \item For instance, as future work we plan to evaluate the performance of the proposed model on evolving graphs of social networks. In particular, provided the limited duration of live video stream events the main challenge resides on identifying the differences of how viewers emerge during live video streaming events and at what pace users establish connections in social networks over time. 
    \item Another interesting future direction is to study the performance of our model on graph snapshots over time steps with different duration. For example, in a live video streaming event the duration of time steps between two consecutive snapshots might vary, depending on the network demand. This means that different time steps might require an adaptive learning strategy of the time window $w$ when training our model.
    \item In our model, training is performed on the graph data of a single live video streaming event. In practice though there are several live video streaming events that take place on a daily basis. The question that we have to answer is how to exploit the knowledge acquired from different live video streaming events, when training our model on a new event. More precisely, we plan to study various transfer learning strategies to exploit the knowledge from different events, when training our model. This is a challenging task, because not only the internal network topologies of several companies vary, but also viewers emerge at various paces during different live video streaming events.
\end{itemize}

\begin{table*}[h]
    \caption{Effect on RMSE when varying the representation size $d$ of the examined models. We report average RMSE over the graph snapshots during the live video streaming event. Bold values denote the best configuration for each model.}
    \resizebox{\textwidth}{!}{
        \centering
        \begin{tabular}{c|c|c|c|c|c|c|c}
        
        \multicolumn{8}{c}{\textbf{LiveStream-4K}}\\ \hline
             \textbf{Representation size} $d$ & \textbf{DynVGAE} & \textbf{EvolveGCN} & \textbf{DySAT} & \textbf{DMTKG}-$\mathcal{T}$ & \textbf{DMTKG}-$\mathcal{S}$ & \textbf{EGAD}-$\mathcal{T}$ & \textbf{EGAD}-$\mathcal{S}$ \\\hline
             16     & $0.23 \pm 0.14$           & $0.27 \pm 0.09$           & $0.18 \pm 0.07$           & $0.28 \pm 0.14$           & $0.27 \pm 0.15$           & $0.17 \pm 0.09$           & $\mathbf{0.13 \pm 0.09}$  \\\hline
             32     & $0.21 \pm 0.12$           & $\mathbf{0.25 \pm 0.12}$  & $0.16 \pm 0.10$           & $0.26 \pm 0.13$           & $\mathbf{0.23 \pm 0.17}$  & $0.16 \pm 0.05$           & $0.15 \pm 0.10$           \\\hline
             64     & $\mathbf{0.19 \pm 0.13}$  & $0.26 \pm 0.18$           & $\mathbf{0.15 \pm 0.09}$  & $\mathbf{0.25 \pm 0.16}$  & $0.24 \pm 0.12$           & $\mathbf{0.14 \pm 0.08}$  & $0.16 \pm 0.06$           \\\hline
             128    & $0.20 \pm 0.17$           & $0.26 \pm 0.12$           & $0.17 \pm 0.08$           & $0.26 \pm 0.17$           & $0.25 \pm 0.16$           & $0.15 \pm 0.06$           & $0.16 \pm 0.08$           \\\hline
             256    & $0.21 \pm 0.15$           & $0.27 \pm 0.15$           & $0.18 \pm 0.05$           & $0.27 \pm 0.19$           & $0.26 \pm 0.14$           & $0.16 \pm 0.07$           & $0.17 \pm 0.06$           \\\hline
                      \multicolumn{8}{c}{\textbf{LiveStream-6K}}\\\hline
             16     & $0.48 \pm 0.16$           & $0.47 \pm 0.12$           &  $0.41 \pm 0.15$          & $0.48 \pm 0.18$           & $0.48 \pm 0.14$           & $0.41 \pm 0.12$           & $\mathbf{0.36 \pm 0.06}$  \\\hline
             32     & $0.46 \pm 0.16$           & $\mathbf{0.44 \pm 0.11}$  & $0.40 \pm 0.17$           & $0.47 \pm 0.18$           & $\mathbf{0.41 \pm 0.12}$  & $0.39 \pm 0.09$           & $0.37 \pm 0.09$           \\\hline
             64     & $\mathbf{0.45 \pm 0.18}$  & $0.45 \pm 0.13$           & $\mathbf{0.39 \pm 0.18}$  & $\mathbf{0.43 \pm 0.16}$  & $0.43 \pm 0.15$           & $\mathbf{0.37 \pm 0.10}$  & $0.39 \pm 0.10$           \\\hline
             128    & $0.46 \pm 0.17$           & $0.45 \pm 0.14$           & $0.41 \pm 0.16$           & $0.45 \pm 0.14$           & $0.44 \pm 0.16$           & $0.38 \pm 0.11$           & $0.40 \pm 0.08$           \\\hline
             256    & $0.46 \pm 0.18$           & $0.46 \pm 0.18$           & $0.42 \pm 0.14$           & $0.46 \pm 0.12$           & $0.46 \pm 0.18$           & $0.38 \pm 0.11$           & $0.42 \pm 0.07$           \\\hline
                      \multicolumn{8}{c}{\textbf{LiveStream-16K}}\\\hline

             16     & $0.36 \pm 0.13$           & $0.35 \pm 0.18$           & $0.29 \pm 0.16$           & $0.39 \pm 0.15$           & $0.39 \pm 0.12$           & $0.27 \pm 0.10$           & $\mathbf{0.23 \pm 0.07}$  \\\hline
             32     & $0.35 \pm 0.14$           & $\mathbf{0.33 \pm 0.19}$  & $0.29 \pm 0.14$           & $0.37 \pm 0.14$           & $0.36 \pm 0.11$           & $0.26 \pm 0.09$           & $0.25 \pm 0.09$           \\\hline
             64     & $\mathbf{0.33 \pm 0.12}$  & $0.34 \pm 0.11$           & $\mathbf{0.27 \pm 0.14}$  & $0.36 \pm 0.15$           & $\mathbf{0.33 \pm 0.11}$  & $\mathbf{0.24 \pm 0.09}$  & $0.26 \pm 0.06$           \\\hline
             128    & $0.34 \pm 0.11$           & $0.35 \pm 0.13$           & $0.28 \pm 0.13$           & $\mathbf{0.35 \pm 0.12}$  & $0.35 \pm 0.15$           & $0.25 \pm 0.10$           & $0.26 \pm 0.08$           \\\hline
             256    & $0.34 \pm 0.17$           & $0.36 \pm 0.11$           & $0.29 \pm 0.12$           & $0.38 \pm 0.13$           & $0.36 \pm 0.18$           & $0.26 \pm 0.10$           & $0.27 \pm 0.10$           \\
        \end{tabular}
    }
    
    \label{tab:repres_4K}
\end{table*}

\begin{table*}[h]
    \caption{Impact of the window size $l$ on the performance of each examined model in terms of RMSE.}
    \resizebox{\textwidth}{!}{
        \centering
        \begin{tabular}{c|c|c|c|c|c|c|c}
        \multicolumn{8}{c}{\textbf{LiveStream-4K}}\\ \hline
             \textbf{Window size} $l$ & \textbf{DynVGAE} & \textbf{EvolveGCN} & \textbf{DySAT} & \textbf{DMTKG}-$\mathcal{T}$ & \textbf{DMTKG}-$\mathcal{S}$ & \textbf{EGAD}-$\mathcal{T}$ & \textbf{EGAD}-$\mathcal{S}$ \\\hline
             1 & $0.32 \pm 0.16$            & $0.35 \pm 0.19$           & $0.21 \pm 0.12$           & N/A & N/A & $0.18 \pm 0.06$           & $0.16 \pm 0.05$ \\\hline
             2 & $\mathbf{0.19 \pm 0.13}$   & $\mathbf{0.25 \pm 0.12}$  & $\mathbf{0.15 \pm 0.09}$  & N/A & N/A & $0.16 \pm 0.09$           & $0.14 \pm 0.06$ \\\hline
             3 & $0.24 \pm 0.18$            & $0.28 \pm 0.14$           & $0.19 \pm 0.14$           & N/A & N/A & $\mathbf{0.14 \pm 0.08}$  & $\mathbf{0.13 \pm 0.09}$ \\\hline
             4 & $0.29 \pm 0.12$            & $0.32 \pm 0.16$           & $0.20 \pm 0.12$           & N/A & N/A & $0.17 \pm 0.10$           & $0.16 \pm 0.04$ \\\hline
             5 & $0.36 \pm 0.17$            & $0.42 \pm 0.19$           & $0.24 \pm 0.14$           & N/A & N/A & $0.20 \pm 0.09$           & $0.19 \pm 0.06$ \\\hline
                     \multicolumn{8}{c}{\textbf{LiveStream-6K}}\\ \hline
             1 & $0.48 \pm 0.16$            & $0.56 \pm 0.11$           & $0.42 \pm 0.14$           & N/A & N/A & $0.41 \pm 0.11$           & $0.39 \pm 0.07$ \\\hline
             2 & $\mathbf{0.45 \pm 0.18}$   & $\mathbf{0.44 \pm 0.11}$  & $\mathbf{0.39 \pm 0.18}$  & N/A & N/A & $0.38 \pm 0.08$           & $0.37 \pm 0.09$ \\\hline
             3 & $0.46 \pm 0.14$            & $0.49 \pm 0.13$           & $0.41 \pm 0.16$           & N/A & N/A & $\mathbf{0.37 \pm 0.10}$  & $\mathbf{0.36 \pm 0.06}$ \\\hline
             4 & $0.52 \pm 0.19$            & $0.51 \pm 0.12$           & $0.43 \pm 0.12$           & N/A & N/A & $0.40 \pm 0.09$           & $0.39 \pm 0.04$ \\\hline
             5 & $0.54 \pm 0.12$            & $0.54 \pm 0.18$           & $0.49 \pm 0.15$           & N/A & N/A & $0.43 \pm 0.12$           &  $0.40 \pm 0.05$\\\hline
                     \multicolumn{8}{c}{\textbf{LiveStream-16K}}\\ \hline
             1 & $0.42 \pm 0.12$            & $0.38 \pm 0.16$           & $0.33 \pm 0.16$           & N/A & N/A & $0.31 \pm 0.10$           & $0.29 \pm 0.06$ \\\hline
             2 & $\mathbf{0.33 \pm 0.12}$   & $\mathbf{0.33 \pm 0.19}$  & $\mathbf{0.27 \pm 0.14}$  & N/A & N/A & $0.29 \pm 0.11$           & $0.27 \pm 0.08$ \\\hline
             3 & $0.37 \pm 0.11$            & $0.36 \pm 0.17$           & $0.29 \pm 0.12$           & N/A & N/A & $\mathbf{0.24 \pm 0.09}$  & $\mathbf{0.23 \pm 0.07}$ \\\hline
             4 & $0.39 \pm 0.18$            & $0.39 \pm 0.19$           & $0.31 \pm 0.16$           & N/A & N/A & $0.30 \pm 0.07$           & $0.29 \pm 0.09$ \\\hline
             5 & $0.46 \pm 0.14$            & $0.41 \pm 0.15$           & $0.39 \pm 0.14$           & N/A & N/A & $0.38 \pm 0.10$           & $0.24 \pm 0.10$ \\

        \end{tabular}
    }
    
    \label{tab:window_4K}
\end{table*}

\begin{table*}[h]
    \caption{Effect on RMSE when varying the number of heads $h$ of the self-attention mechanisms of DySAT, EGAD-$\mathcal{T}$ and EGAD-$\mathcal{S}$.}
    \resizebox{\textwidth}{!}{
        \centering
        \begin{tabular}{c|c|c|c|c|c|c|c}
         \multicolumn{8}{c}{\textbf{LiveStream-4K}}\\ \hline
             \textbf{Number of heads} $h$ & \textbf{DynVGAE} & \textbf{EvolveGCN} & \textbf{DySAT} & \textbf{DMTKG}-$\mathcal{T}$ & \textbf{DMTKG}-$\mathcal{S}$ & \textbf{EGAD}-$\mathcal{T}$ & \textbf{EGAD}-$\mathcal{S}$ \\\hline
             1 & N/A & N/A & $0.19 \pm 0.06$            & N/A & N/A & $0.16 \pm 0.05$           & $\mathbf{0.13 \pm 0.09}$  \\\hline
             2 & N/A & N/A & $0.17 \pm 0.07$            & N/A & N/A & $0.15 \pm 0.04$           & $0.14 \pm 0.10$           \\\hline
             3 & N/A & N/A & $\mathbf{0.15 \pm 0.09}$   & N/A & N/A & $\mathbf{0.14 \pm 0.08}$  &  $0.16 \pm 0.08$          \\\hline
             4 & N/A & N/A & $0.16 \pm 0.08$            & N/A & N/A & $0.16 \pm 0.07$           & $0.17 \pm 0.09$           \\\hline
             5 & N/A & N/A & $0.18 \pm 0.08$            & N/A & N/A & $0.17 \pm 0.09$           & $0.20 \pm 0.03$           \\\hline
                      \multicolumn{8}{c}{\textbf{LiveStream-6K}}\\ \hline
             1 & N/A & N/A & $0.47 \pm 0.12$            & N/A & N/A & $0.42 \pm 0.07$           & $\mathbf{0.36 \pm 0.06}$ \\\hline
             2 & N/A & N/A & $0.45 \pm 0.15$            & N/A & N/A & $0.40 \pm 0.12$           & $0.38 \pm 0.06$ \\\hline
             3 & N/A & N/A & $\mathbf{0.39 \pm 0.18}$   & N/A & N/A & $\mathbf{0.37 \pm 0.10}$  & $0.41 \pm 0.04$ \\\hline
             4 & N/A & N/A & $0.41 \pm 0.14$            & N/A & N/A & $0.38 \pm 0.10$           & $0.43 \pm 0.09$ \\\hline
             5 & N/A & N/A & $0.46 \pm 0.17$            & N/A & N/A & $0.39 \pm 0.11$           & $0.43 \pm 0.08$ \\\hline
             \multicolumn{8}{c}{\textbf{LiveStream-16K}}\\ \hline
             1 & N/A & N/A & $0.32 \pm 0.17$            & N/A & N/A & $0.28 \pm 0.08$           & $\mathbf{0.23 \pm 0.07}$   \\\hline
             2 & N/A & N/A & $0.29 \pm 0.18$            & N/A & N/A & $0.26 \pm 0.10$           & $0.25 \pm 0.05$            \\\hline
             3 & N/A & N/A & $0.28 \pm 0.15$            & N/A & N/A & $\mathbf{0.24 \pm 0.09}$  & $0.26 \pm 0.09$            \\\hline
             4 & N/A & N/A & $\mathbf{0.27 \pm 0.14}$   & N/A & N/A & $0.25 \pm 0.06$           &  $0.26 \pm 0.10$          \\\hline
             5 & N/A & N/A & $0.29 \pm 0.12$            & N/A & N/A & $0.27 \pm 0.07$           & $0.27 \pm 0.08$             \\             
        \end{tabular}
    }
    
    \label{tab:heads_4K}
\end{table*}

\bibliographystyle{IEEEtran}
\bibliography{IEEEexample}

\end{document}